\documentclass[runningheads]{llncs}

\usepackage{eccv}  

\usepackage{eccvabbrv}
\usepackage{graphicx}
\usepackage[accsupp]{axessibility}  
\usepackage{capt-of,xcolor,xspace,enumitem}
\usepackage{amsmath,amssymb,amsbsy,amsfonts,dsfont,pifont,bm,bbm,mathrsfs,mathtools,nicefrac}
\usepackage{algorithm,algpseudocode,listings}
\usepackage{booktabs,multirow,adjustbox,diagbox,threeparttable,tabularray}

\usepackage[pagebackref,breaklinks,colorlinks,citecolor=eccvblue]{hyperref}

\usepackage{orcidlink}
\usepackage{wrapfig}
\usepackage[capitalize]{cleveref}  
\crefname{section}{Sec.}{Secs.}
\Crefname{section}{Section}{Sections}
\crefname{table}{Tab.}{Tabs.}
\Crefname{table}{Table}{Tables}
\crefname{figure}{Fig.}{Figs.}
\Crefname{figure}{Figure}{Figures}
\crefname{equation}{Eq.}{Eqs.}
\Crefname{equation}{Equation}{Equations}
\usepackage{titletoc}

\newcommand{\tocite}[1]{{\color{red} [TO CITE]}}
\newcommand{\method}{\textcolor{black}{\texttt{DreamLIP}}\xspace}

\usepackage{tablefootnote}
\usepackage{hyperref}

\colorlet{darkgreen}{green!65!black}
\colorlet{darkblue}{blue!75!black}
\colorlet{darkred}{red!80!black}
\definecolor{lightblue}{HTML}{0071bc}
\definecolor{lightgreen}{HTML}{39b54a}

\usepackage{amsmath,amsfonts,bm}

\def\eqref#1{equation~\ref{#1}}

\def\1{\bm{1}}

\def\vtheta{{\bm{\theta}}}

\def\vt{{\bm{t}}}

\def\vv{{\bm{v}}}

\def\mV{{\bm{V}}}
\def\mW{{\bm{W}}}

\DeclareMathAlphabet{\mathsfit}{\encodingdefault}{\sfdefault}{m}{sl}
\SetMathAlphabet{\mathsfit}{bold}{\encodingdefault}{\sfdefault}{bx}{n}

\def\gL{{\mathcal{L}}}

\begin{document}

\title{DreamLIP: Language-Image Pre-training with Long Captions} 
\titlerunning{DreamLIP}

\author{
    Kecheng Zheng\thanks{equal contribution}\inst{,1,2} \and
    Yifei Zhang\inst{\star,3} \and
    Wei Wu\inst{4} \and
    Fan Lu\inst{4} \and  \\
    Shuailei Ma\inst{6} \and
    Xin Jin\inst{5} \and
    Wei Chen \inst{1}\and
    Yujun Shen \inst{2}
}
\authorrunning{Z.~Author et al.}

\institute{
    State Key Lab of CAD\&CG, Zhejiang University \and
    \mbox{Ant Group      \and
    Shanghai Jiao Tong University} \and
    University of Science and Technology of China \and
    \mbox{Eastern Institute of Technology       \and
    Northeastern University, China}\\
    \email{zkechengzk@gmail.com}
}

\maketitle

\begin{abstract}

Language-image pre-training largely relies on how precisely and thoroughly a text describes its paired image.
In practice, however, the contents of an image can be so rich that well describing them requires lengthy captions (\textit{e.g.}, with 10 sentences), which are usually missing in existing datasets.
Consequently, there are currently no clear evidences on \textit{whether and how language-image pre-training could benefit from long captions}.
To figure this out, we first re-caption 30M images with detailed descriptions using a pre-trained Multi-modality Large Language Model (MLLM), and then study the usage of the resulting captions under a contrastive learning framework.
We observe that, each sentence within a long caption is very likely to describe the image partially (\textit{e.g.}, an object).
Motivated by this, we propose to dynamically sample sub-captions from the text label to construct multiple positive pairs, and introduce a grouping loss to match the embeddings of each sub-caption with its corresponding local image patches in a self-supervised manner.
Experimental results on a wide rage of downstream tasks demonstrate the consistent superiority of our method, termed \method, over previous alternatives, highlighting its fine-grained representational capacity.
It is noteworthy that, on the tasks of image-text retrieval and semantic segmentation, our model trained with 30M image-text pairs achieves on par or even better performance than CLIP trained with 400M pairs.
Project page is available at~\url{https://zyf0619sjtu.github.io/dream-lip}.


\end{abstract}

\section{Introduction}\label{sec:intro}
\begin{figure*}[t]
    \centering
    \includegraphics[width=0.98\linewidth]{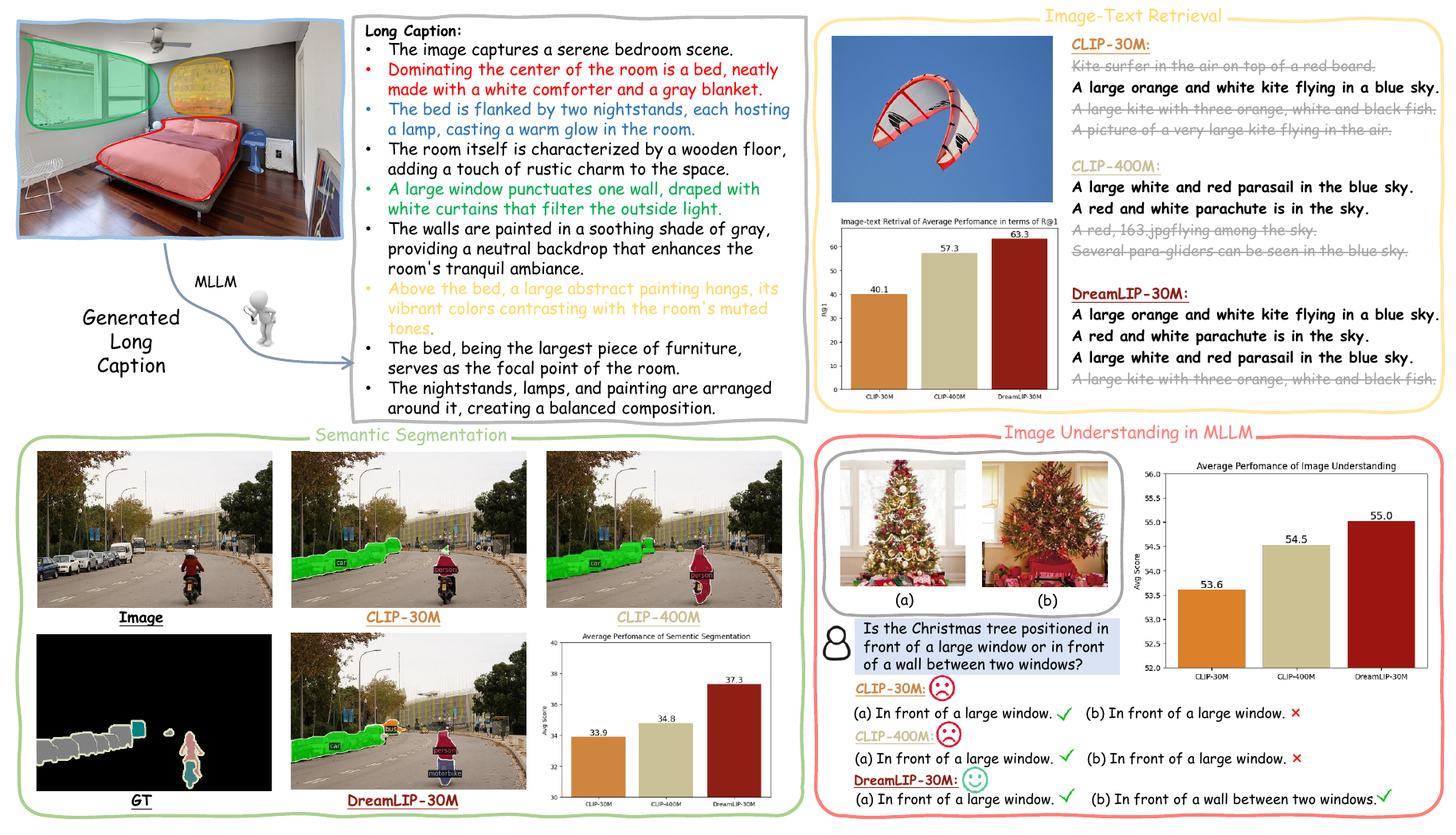}
    \caption{%
        The richness of an image's content often necessitates long captions for adequate description, with each sentence likely conveying a fragment of the image's entirety. Thanks to the long captions, our \method trained with 30M image-text pairs achieves on par or even better performance than CLIP trained with 400M pairs on the tasks of image-text retrieval, semantic segmentation, and image understanding in MLLM.
    }
    \label{fig:moti}
\end{figure*}

Language-image pre-training largely relies on how precisely and thoroughly a text describes its paired image.
In the existing image-text paired datasets, an image is described by a short caption, which barely scratches the surface of the intricate real image. 
In practice, the contents of an image can be so rich that describing them well requires long captions (\textit{e.g.}, with 10 sentences).
These sentences are included in the long captions of corresponding image which can typically delineate various local regions of the image.
Thus, long captions can unleash the potential of a real-world image and enrich semantic learning of language-image models.
This multifaceted relationship between long captions and images presents an untapped reservoir of semantic richness.

Although long captions have enormous potential, there is a shortage of million-scale datasets to evaluate it in vision-language pre-training.
One of the most straightforward approaches is: utilizing pre-trained Multi-modality Large Language Models (MLLM) to generate long captions.
Actually, recent works \cite{zhao2023rleg,dalle3,wu2024imagine,tian2024stablerep,fan2024improving,yang2023alip} also have explored the use of synthetic data to improve vision-language modeling.
Specifically, RLEG~\cite{zhao2023rleg} exploits DALL-E 2~\cite{ramesh2022hierarchical} to generate feature embedding online for learning effective vision-language representation.
StableRep~\cite{tian2024stablerep} shows that training on synthetic images generated by Stable Diffusion~\cite{Rombach_2022_CVPR} and their multi-positive contrastive learning method can match the performance of training on real images.
LaCLIP~\cite{fan2024improving} uses large language models to rewrite the captions of the real images as data augmentation.
On the other hand, some methods~\cite{yang2023alip, liu2023mllms} utilize the pre-trained visual-language model to generate concise synthetic \textit{short captions} that focus on the image content to address the noise in the original web data.
%
%
Although existing methods have attempted to improve the representation ability of multi-modal models using short captions, 
there are currently no clear evidences on \textit{whether and how language-image pre-training could benefit from long captions}.

To figure this out, we first re-caption 30M images with detailed descriptions using a pre-trained MLLM, and then study the usage of the resulting captions under a contrastive learning framework.
As shown in~\cref{fig:moti}, we observe that each sentence within a long caption is very likely to describe the image partially (\textit{e.g.}, bed).
Motivated by this, we propose to dynamically sample sub-captions from the text label to construct multiple positive pairs.
This innovative process allows us to forge an image-text dataset with long captions, wherein we employ a multi-positive loss framework to intricately intertwine sub-captions with their corresponding images, crafting a rich tapestry of aligned modalities. 
Taking it a step further, our approach also delves into learning subcaption-specific groupings of image patches. 
By applying a subcaption-specific grouping loss, we achieve a fine-grained alignment that meticulously pairs local image patches with their respective textual embeddings across different sub-captions. 
This nuanced pairing brings forth a better semantic alignment between the two modalities. 

Experimental results across a diverse array of downstream tasks consistently demonstrate the superiority of our \method. 
These results underscore its exceptional fine-grained representational abilities when compared to previous alternatives. 
Notably, in the tasks of image-text retrieval, semantic segmentation, and image understanding in MLLM, our model trained on 30 million image-text pairs datasets achieves performance that is comparable to, or even surpasses, that of CLIP, despite the latter being trained on a dataset consisting of 400 million pairs.

\section{Related Work}\label{sec:related}
\noindent\textbf{Vision-Language Pre-training.}
Some recent works like CLIP\cite{radford2021learning} and ALIGN\cite{jia2021scaling} have shown that contrastive vision-language pre-training can provide rich and general representations for numerous downstream tasks.
However, these methods only apply contrastive loss between the entire image and its caption, ignoring the local alignment between text and image.
To this end, FILIP\cite{yao2021filip} and PyramidCLIP\cite{gao2022pyramidclip} modified the original contrastive loss to act between text tokens and image patches.
Furthermore, HILCLIP\cite{geng2023hiclip} captures the hierarchical nature of high-level and fine-grained semantics conveyed in images and texts through hierarchy-aware attention.
UniCLIP\cite{lee2022uniclip} instead integrated the contrastive loss of both inter-domain and intra-domain pairs into a single universal space.
SoftCLIP\cite{gao2023softclip} relaxed the strict one-to-one constraint with a soft target to enable image-text pairs to have some local similarities and model many-to-many relationships between the two modalities.
Instead, in order to get a more robust representation, CLOOB\cite{furst2022cloob} used Hopfield network to regulate the convergence of the learned representations, and FIBER\cite{dou2022coarse} introduced dual modality encoders to obtain better fused multi-modal features.
In addition, FLIP\cite{li2023scaling} and MaskCLIP\cite{dong2023maskclip} used image patch masking as an effective method for vision-language pre-training.
Finally, since the image-text pairs crawled from the internet are replete with substantial noise, UniCL\cite{yang2022unified} proposed to integrate image-label pairs from supervised datasets, while CoCa\cite{yu2022coca}, BLIP\cite{li2022blip} proposed to denoise text captions by regenerating them.
%
%

%
\noindent\textbf{Improving Vision-Language Pre-training with Synthetic Data.}
Synthetic data has been employed to improve models' performance on many computer vision tasks ranging from semantic segmentation\cite{richter2016playing, chen2019learning}, object detection\cite{johnson2016driving} and image classification\cite{yuan2023real}.
Recent works \cite{zhao2023rleg,tian2024stablerep,fan2024improving,yang2023alip} have explored the effect of synthetic data to improve vision-language pre-training.
RLEG\cite{zhao2023rleg} exploits DALL-E 2\cite{ramesh2022hierarchical} to generate feature embedding online for learning effective vision-language representation.
StableRep\cite{tian2024stablerep} shows that training on synthetic images generated by Stable Diffusion\cite{Rombach_2022_CVPR} and their multi-positive contrastive learning method can match the performance of training on real images.
Instead, SynthCLIP\cite{hammoud2024synthclip} creates synthetic image-text data using text-to-image generative networks and large language models to eliminate the impact of misalignment, long-tail distribution, and harmful contents in real data.
On the other hand, ALIP\cite{yang2023alip} utilizes the OFA model to generate correct synthetic captions that focus on the image content to address the noise in the original web data.
Similarly, LaCLIP\cite{fan2024improving} uses large language models to rewrite the captions of the real images as data augmentation.
The closest to our work is the concurrent new works \cite{liu2023mllms,veclip} that enhance visual-language representation learning by utilizing multi-modal large language models to generate captions for each image.
Although existing methods have attempted to improve the representation ability of multi-modality models using short captions, they have yet to investigate the employment of \textit{long captions}, which are rich in image details and offer untapped informational potential.
\section{Method}\label{sec:method}

\begin{figure}[t]
    \centering
    \includegraphics[width=0.98\linewidth]{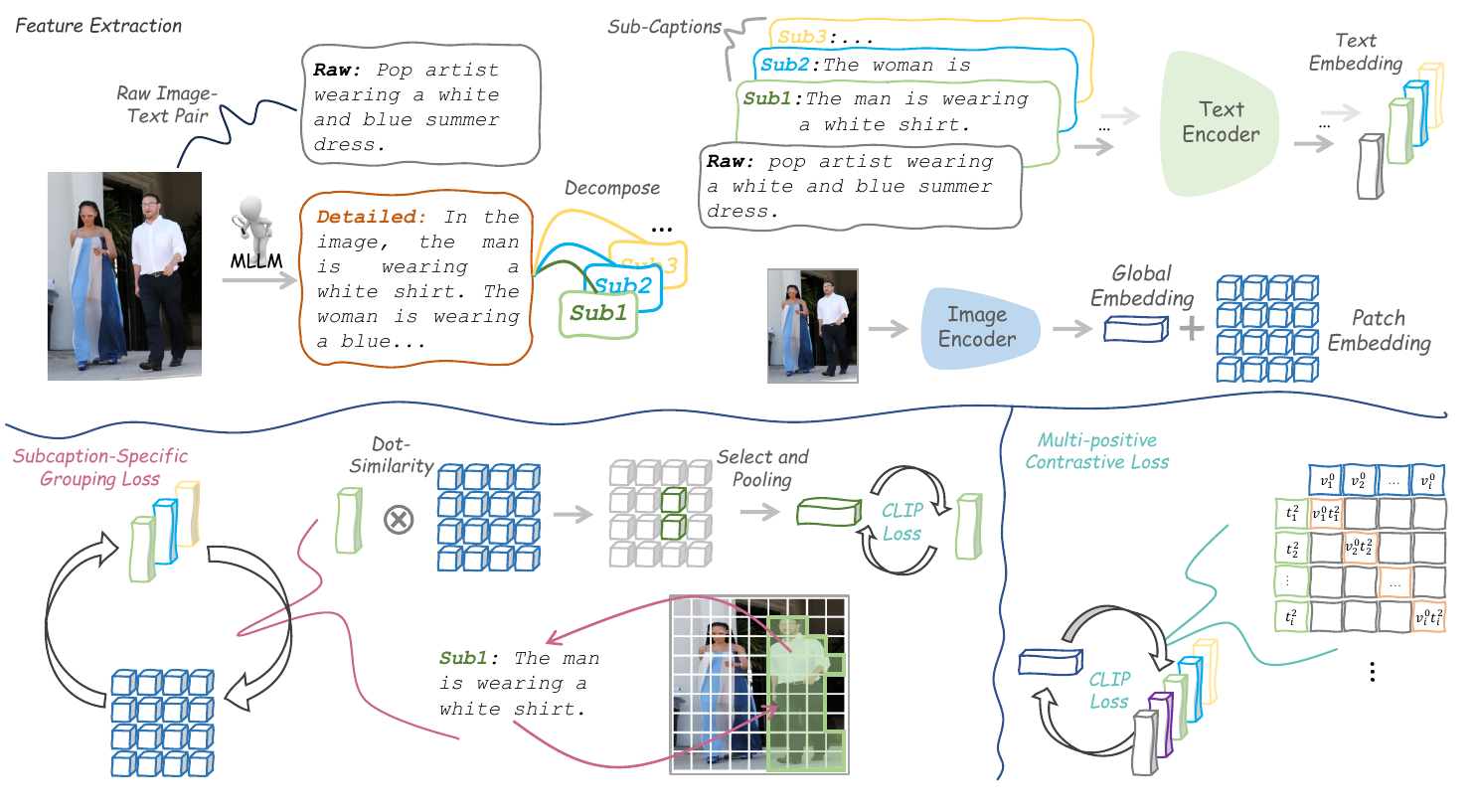}
    \vspace{-2mm}
    \caption{
        Illustration of \method. 
        Firstly, we dynamically sample sub-captions from the text label to construct multiple positive pairs.
        Then, a global multi-positive contrastive loss is used to align text embeddings of sub-captions and global image embedding.
        Meanwhile, we introduce a grouping loss to match the embeddings of each sub-caption with its corresponding local image patches in a self-supervised manner.
        }
    \label{fig:arch}
    \vspace{-5mm}
\end{figure}

\subsection{Preliminary}
\label{sec:method_clip}
CLIP~\cite{radford2021learning} mainly consists of two components: an image encoder $E_I$ and a text encoder $E_T$, which can project image and text into the same feature embedding space. 
Specifically, the images $\{I_1, I_2,\cdots,I_N\}$ and the corresponding raw short texts $\{T_1, T_2,\cdots,T_N\}$ are sampled from the training dataset during each training step. 
The features of image $I_i$ and text $T_i$ are extracted using dedicated encoders and normalization functions: $\boldsymbol{v}_{i} = E_I(I_i;\vtheta), \boldsymbol{t}_i = E_T(T_i;\bm{\beta}).$
The image and text features are used to compute the InfoNCE\cite{oord2018representation} loss, where the paired image-text forms the positive pairs, and the unpaired ones are treated as negative samples.
In this way, the text-to-vision loss can be obtained by:
%
\begin{gather}
    \gL^{t2v}=-\sum_{i=1}^N \log \frac{\exp \left(\cos \langle \boldsymbol{v}_i, \boldsymbol{t}_i\rangle / \tau\right)}{\sum_{j=1}^N \exp \left(\cos \langle\boldsymbol{v}_j, \boldsymbol{t}_i\rangle / \tau\right)},
    \label{eq.t2v_loss}
\end{gather}
where $\text{cos}\langle \cdot, \cdot \rangle$ denotes the cosine similarity and $\tau$ is a learnable temperature parameter.
Similarly, the vision-to-text loss can be obtained by:
\begin{gather}
    \gL^{v2t}=-\sum_{i=1}^N \log \frac{\exp \left(\cos \langle \boldsymbol{t}_i, \boldsymbol{v}_i\rangle / \tau\right)}{\sum_{j=1}^N \exp \left(\cos \langle\boldsymbol{t}_j, \boldsymbol{v}_i\rangle / \tau\right)}.
    \label{eq.v2t_loss}
\end{gather}
%
%
%
The total training loss is $\gL_{\text{CLIP}} =( \gL^{t2v}$ + $\gL^{v2t}) / 2$. 

\subsection{Synthetic Long Caption from Image}

The utilization of paired image-text datasets is pivotal in fostering models capable of perceiving semantically aligned information across modalities.
However, most of these datasets include a non-negligible portion of noisy and mismatched image-text pairs, substantially affecting visual-language representation learning.
Thus, existing methods apply some short captions generated from MLLMs for real images to improve representation learning of language-vision pretrained models.
The existing approaches, which correlate each image with a short caption, barely scratch the surface of the intricate tapestry woven by real-world data. 
They ignore exploring how to unleash the potential of long captions that can describe a real image in detail.

Given an image-text paired dataset $\mathcal{D}=\{(I_i,T_i)\}_{i=1}^{N}$, containing $N$ image-text pairs, we adopt a pretrained MLLM as a captioner $f$ to generate a collection of long and short captions of images:
\begin{equation}
    \mathcal{C} = \{C^l_i, C^s_i\}^{N}_{i=1}=\{f(I_i,q_l), f(I_i,q_s)\}^{N}_{i=1},
\label{eq:multicaption}
\end{equation}
where $C^l_i$ and $C^s_i$ denote the long caption and the short caption of image $I_i$.
In addition, $q_s$ and $q_l$ denote the text inputs for generating long and short captions to the MLLM, respectively.
We use the simple question template \textit{``Describe the image in details''} to query the detailed long captions. 
Meanwhile, \textit{``Describe the image in short:''} is used as the prompt for generating short captions.
The question of detailed long captions has little impact on the diversity of answers, so we can obtain comprehensive captions of each image. 
Short captions, being concise and less prone to inaccuracies, naturally complement their longer captions by providing a succinct essence of the content.
\textbf{How long are these long captions?} For most raw/short caption labels, the number of captions in the label is one or two, with a token count of approximately 20. 
An image can include much effective information that needs substantial captions to describe its visual content.
Thus, We have counted the number of tokens and sub-captions to see the amount of information in the generated long captions.
Although there are some hallucinations in the generated caption, long captions can still bring much effective information that accurately describes the image. 
%
%
%
We also use different MLLMs (\eg, InstructBLIP and LLaVA-1.5) to generate long captions to help CLIP model training. 
The experimental results and statistics of different long captions are presented in~\cref{sec:dMLLM}.

\subsection{Global Multi-Positive Contrastive Learning}

Sometimes complex and multiple information can be conveyed by a single image, which conveys its meaning or essence more effectively than a mere verbal description.
Thus, a picture, which is worth a thousand words, should be described in multiple sentences.
Inspired by this, we would like to design a strategy for long captions, towards completely utilizing the information of an image.
Given original caption $T$, generated long caption $C^l = [c_1, \dots, c_M]$ and short caption $C^s=[c_s]$, we firstly conduct a sub-caption set that includes different sub-captions. 
Then we can easily implement a straightforward random sampling process to sample some sub-captions from the sub-caption set:
\begin{equation}
    S_{i,j} \sim \text{Uniform}([T, c_s, c_1, \dots, c_M]),
\end{equation}
where $S_{i,j}$ refers to $j$-th sub-caption of $i$-th sample from the sub-caption set. Following LaCLIP~\cite{fan2024improving}, the training multi-positive loss over the images becomes:
\begin{equation}
    \gL_{\text{MPCL}}^{t2v}=-\sum_{i=1}^N \sum_{j=1}^{K} \log \frac{\exp \left(\cos \langle \vv_i, \vt_{i,j}\rangle / \tau\right)}{\sum_{n=1}^N \exp \left(\cos \langle \vv_n, \vt_{i,j}\rangle / \tau\right)},
\end{equation}
where $\vt_{i,j}$ refers to the text embedding of sub-caption $S_{i,j}$, and $K$ denotes the number of sampling sub-captions. Similar to CLIP loss, $\gL_{\text{MPCL}} =( \gL_{\text{MPCL}}^{t2v}$ + $\gL_{\text{MPCL}}^{v2t}) / 2$.
This innovative process allows us to forge an image-text dataset with long captions, wherein we employ a multi-positive loss framework to intricately intertwine sub-captions with their corresponding images, crafting a rich tapestry of aligned modalities. 

\subsection{Subcaption-specific Grouping Loss}
Multi-positive contrastive learning with long captions can help the model perceive better global representation.
To enhance the fine-grained representation of CLIP, existing methods~\cite{yao2021filip} try to utilize word tokens to align the image patches, which may introduce the alignment of words unrelated to visual concepts (\eg, emotion words or conjunctions).
Meanwhile, words in a short caption may not take care of whole image.
Thus, we would like to delve into learning subcaption-specific groupings of image patches for improving the fine-grained ability of the vision-language pre-training model.
Multiple sub-captions from the long caption can typically delineate various local regions of the image.
Given an image and its long caption, we first calculate the cosine similarity between text embedding $\vt_{i}$ of $i$-th sub-caption and its image patch embedding $\mV = [ \vv_1, \ldots, \vv_{HW} ]$ to localize the subcaption-specific groupings of image patches.
%
%
We view the cross-attention weights as a similarity matrix $\mW = \{\hat{w}_{i,j}\}$ and sparse it to let each sub-caption only focus on a few visual tokens, \ie
\begin{equation}
  \tilde{w}_{i,j} =
    \begin{cases}
      \hat{w}_{i,j} & \text{if $\hat{w}_{i,j} \geq \sigma$} \\
      0 & \text{otherwise} \\
    \end{cases}   
\end{equation}
where $\sigma$ is the sparsity threshold. 
After that, these weights are used to select some subcaption-specific grouping visual tokens and pool them together:
\begin{equation}
    \hat{\vv}_j = \sum_{n=1}^{HW} \frac{\tilde{w}_{i,j}}{\sum_{j} \tilde{w}_{i,j}} \bm{v}_n.
\end{equation}
The subcaption-specific grouping loss can be formulated as follows
\begin{equation}
    \gL_\text{Sub}=-\sum_{i=1}^N \sum_{j=1}^{M+2} \log \frac{\exp \left(\cos \left(\hat{\vv}_{i,j}, \vt_{i,j}\right) / \tau\right)}{\sum_{n=1}^{K} \exp \left(\cos \left(\hat{\vv}_{i,n}, \vt_{i,j}\right) / \tau\right)},
\end{equation}

By applying this subcaption-specific grouping loss, we achieve a fine-grained alignment that meticulously pairs local image patches with their respective textual embeddings across different sub-captions. 
This nuanced pairing brings forth an unprecedented level of semantic alignment between the two modalities.

\subsection{Overall objective} 
The overall \method objective is a weighted sum of the multi-positive contrastive loss and the finegrained alignment constrastive loss: 
\begin{equation}
    \gL_{\text{\method}} = \lambda_{MPCL} \gL_{\text{MPCL}} + \lambda_S L_\text{Sub},
\end{equation}
where the loss weights $\lambda_{MPCL}$ and $\lambda_S$ are hyperparameters.
\section{Experiments}\label{sec:exp}

\subsection{Implementation Details and Datasets}
\noindent \textbf{Pretraining Datasets.}
To make a fair comparison with the state-of-the-art contrastive vision-language pretraining approaches, we adopt the CC3M, CC12M and YFCC15M datasets. In addition, we construct a \textbf{30M} version of pretraining data by including Conceptual Caption 3M (CC3M)~\cite{sharma-etal-2018-conceptual} and 12M (CC12M)~\cite{sharma-etal-2018-conceptual}. We mainly conduct ablation studies to validate our model on the CC3M data.

\begin{table*}[t!]
    \caption{Zero-shot image-text retrieval on the test splits of Flickr30k and MSCOCO. All models are pre-trained on YFCC15M. We use ViT-B/32 as image backbone. * denotes that we report results for pre-trained ViT-B/32 from OpenCLIP code base. Long caption generated from ShareGPT4V~\cite{sharegpt4v} is used.}
    \label{tab:retrieval}

    \centering\scriptsize
    \SetTblrInner{rowsep=1.2pt}      
    \SetTblrInner{colsep=1.4pt}      
    \resizebox{\linewidth}{!}{
    \begin{tblr}{
        cells={halign=c,valign=m},   
        column{1}={halign=l},        
        hline{4,6,8,18,20}={1-14}{}, 
        hline{1,21}={1.0pt},         
        cell{1}{1}={r=3}{},          
        cell{1}{2}={r=3}{},          
        cell{4}{1}={r=2}{},          
        cell{6}{1}={r=2}{},          
        cell{8}{1}={r=10}{},          
        cell{18}{1}={r=2}{},         
        cell{1}{3}={c=6}{},          
        cell{1}{9}={c=6}{},          
        cell{2}{3}={c=3}{},          
        cell{2}{6}={c=3}{},          
        cell{2}{9}={c=3}{},          
        cell{2}{12}={c=3}{},         
    }
        Data       & Method & Text Retrieval &     &      &        &     &      & Image Retrieval &     &      &        &     &      \\
                   &        & Flickr30k      &     &      & MSCOCO &     &      & Flickr30k       &     &      & MSCOCO &     &      \\
                   &        & R@1            & R@5 & R@10 & R@1    & R@5 & R@10 & R@1             & R@5 & R@10 & R@1    & R@5 & R@10 \\
        CC3M       & CLIP\cite{radford2021learning} &26.6 & 52.5 & 63.2 & 13.4 & 32.0 & 43.3 & 18.3 & 39.4 & 49.7 & 10.1 & 25.6 & 35.7\\
                   & \method                        &\bf 57.6 & \bf84.4 &\bf 89.6 &\bf 33.4 & \bf60.7 &\bf 72.0 &\bf 42.2 &\bf 69.0 &\bf 77.7 &\bf 23.4 &\bf 47.2 &\bf 58.6 \\
        CC12M      & CLIP\cite{radford2021learning} &49.3 & 77.3 & 85.0 & 29.3 & 54.4 & 65.3 & 35.5 & 61.8 & 71.6 & 19.0 & 41.0 & 52.5\\
                   & \method                        &\bf 78.7 &\bf 94.6 &\bf 97.6 & \bf53.4 & \bf77.1 &\bf 84.7 &\bf 61.0 &\bf 83.9 &\bf 89.8 &\bf 36.7 & \bf62.3 &\bf 72.3 \\
        YFCC15M    & CLIP\cite{radford2021learning} & 34.9 & 63.9 & 75.9 & 20.8 & 43.9 & 55.7 & 23.4 & 47.2 & 58.9 & 13.0 & 31.7 & 42.7 \\
                   & SLIP\cite{mu2022slip}          & 47.8 & 76.5 & 85.9 & 27.7 & 52.6 & 63.9 & 32.3 & 58.7 & 68.8 & 18.2 & 39.2 & 51.0 \\
                   & DeCLIP\cite{li2021supervision} & 51.4 & 80.2 & 88.9 & 28.3 & 53.2 & 64.5 & 34.3 & 60.3 & 70.7 & 18.4 & 39.6 & 51.4 \\
                   & UniCLIP\cite{lee2022uniclip}   & 52.3 & 81.6 & 89.0 & 32.0 & 57.7 & 69.2 & 34.8 & 62.0 & 72.0 & 20.2 & 43.2 & 54.4 \\
                   & MCD\cite{kim2023misalign}                            & 57.6 & 82.6 & 91.1 & 32.3 & 58.7 & 71.2 & 36.4 & 64.8 & 74.1 & 20.7 & 43.5 & 55.3 \\
                   & HiCLIP\cite{geng2023hiclip}    & -    & -    & -    & 34.2 & 60.3 & 70.9 & -    & -    & -    & 20.6 & 43.8 & 55.3 \\
                   & HiDeCLIP\cite{geng2023hiclip}  & -    & -    & -    & 38.7 & 64.4 & 74.8 & -    & -    & -    & 23.9 & 48.2 & 60.1 \\
                   & FILIP\cite{yao2021filip}                           &-&-&-&33.4 &60.1&-&-&-&-&23.0 &46.2&- \\
                   & ALIP\cite{yang2023alip}                           & 70.5 & 91.9 & 95.7 & 46.8 & 72.4 & 81.8 & 48.9 & 75.1 & 82.9 & 29.3 & 54.4 & 65.4 \\
                   & \method & \bf 84.9 & \bf 97.3 & \bf 98.7 & \bf 55.7 & \bf 80.5 & \bf 88.2 & \bf 66.0 & \bf 86.4 & \bf 91.4 & \bf 39.8 & \bf 66.0 & \bf 75.5 \\
        Merged-30M & CLIP~\cite{radford2021learning}                          & 57.8 & 84.6 & 91.7 & 35.0 & 61.9 & 73.7 & 44.0 & 70.9 & 79.7 & 23.5 & 47.3 & 59.1\\
                   & \method & \bf 87.2 & \bf 97.5 & \bf 98.8 & \bf 58.3 & \bf 81.6 & \bf 88.8 & \bf 66.4 & \bf 88.3 & \bf 93.3 & \bf 41.1 & \bf 67.0 & \bf 76.6\\
        LAION-400M* & CLIP~\cite{radford2021learning}                           & 78.7 & 94.0 & 97.3 & 53.7 & 77.1 & 85.4 & 61.8 & 85.5 & 90.8 & 34.8 & 60.4 & 71.1\\
    \end{tblr}}
    \vspace{-10pt}
\end{table*}

\noindent \textbf{Downstream Datasets.}
Following CLIP, we select 11 visual recognition datasets under the zero-shot setting, namely ImageNet~\cite{deng2009imagenet}, CIFAR 10 \& CIFAR 100~\cite{krizhevsky2009learning}, StanfordCars~\cite{KrauseStarkDengFei-Fei_3DRR2013}, Caltech101~\cite{fei2004learning}, Flowers102~\cite{nilsback2008automated}, SUN397~\cite{xiao2010sun}, DTD~\cite{cimpoi2014describing}, FGVCAircraft~\cite{maji2013fine}, OxfordPets~\cite{parkhi2012cats}, and Food101~\cite{bossard2014food}. 
The same zero-shot classification protocol is applied following \cite{radford2021learning}, which uses predefined prompts as text inputs. Although CLIP only evaluates on visual recognition, we also provide comprehensive comparisons on vision-language tasks which are more desired in evaluating multi-modal models, including image-text retrieval on MSCOCO/Flickr30k Caption~\cite{lin2014microsoft,young2014image}, semantic segmentation (\ie, ADE20K-150~\cite{zhou2017scene}, ADE20K-847~\cite{zhou2017scene}, VOC-20~\cite{everingham2012pascal}, PC-59~\cite{mottaghi2014role} and PC-459~\cite{mottaghi2014role}) as well as vision-language reasoning on ScienceQA-IMG \cite{sqa}, TextVQA \cite{Textqa}, POPE \cite{pope} and MMVP \cite{tong2024eyes}.

\noindent \textbf{Implementation Details.}
Two variants of Vision Transformer are used as the image encoder in our experiments, \ie, ViT-B/32 and ViT-B/16, while the text encoder is a vanilla Transformer following CLIP as a fair comparison. The embedding size of both image and text features are $512$ throughout our paper. To make a fair comparison with CLIP family baselines, we train all models for $32$ epochs under the same set of pretraining hyperparameters including learning rate, warmup steps, weight decay, \textit{etc.} The input image size is set to $224 \times 224$, and the input text sequence length is truncated or padded to $77$. Following CLIP, the learnable temperature parameter $\tau$ is initialized as $0.07$.

\subsection{Image-Text retrieval}

After pre-training, the proposed model is evaluated in a zero-shot setting on image-text retrieval tasks, \ie, COCO, and Flickr30K. 
The pre-trained model is applied to extract embeddings from images and texts, respectively.
Similarity scores between image embeddings and text embeddings are used for ranking. We use the R@K to report the recall of top-K retrieval items.
As shown in~\Cref{tab:retrieval}, quantitative experimental results demonstrate our superiority over state-of-the-art alternatives in terms of all metrics. 
Thanks to long captions, our model can achieve better performance than CLIP trained by 400M image-text datasets.

\begin{table*}[t]
    \caption{Transferable performance of semantic segmentation on ADE-847, PC-459, ADE-150, PC-59, and VOC-20. Following SAN~\cite{xu2023side}, we used the full training set of COCO-stuff as the training data and our \method as pretrained models. We use ViT-B/32 as image backbone. Long caption generated from ShareGPT4V~\cite{sharegpt4v} is used.}
    \label{tab:ss}
    \centering\scriptsize
    \SetTblrInner{rowsep=1.2pt}      
    \SetTblrInner{colsep=6.8pt}      
    \resizebox{\linewidth}{!}{
    \begin{tblr}{
        cells={halign=c,valign=m},   
        column{1}={halign=l},        
        hline{2,4,6,8,10}={1-8}{},           
        hline{1,11}={1-8}{1.0pt},         
        vline{2,3,4,5,6,7,8}={1-21}{},      
        cell{2}{1}={r=2}{},          
        cell{4}{1}={r=2}{},          
        cell{6}{1}={r=2}{},          
        cell{8}{1}={r=2}{},          
    }
        Data & Method & ADE-847& PC-459 &ADE-150 &PC-59& VOC-20&avg. \\
        CC3M       & CLIP~\cite{radford2021learning} &2.1 & 5.2 & 12.3 & 33.8 & 65.4 &23.8\\
                   & \method  &4.1 & 7.5 & 17.1 & 39.9 & 76.5&29.0 \\
        CC12M      & CLIP~\cite{radford2021learning} &3.3 & 6.7 & 15.7 & 39.2 & 79.7&28.9 \\
                   & \method  &6.1 & 10.0 & 23.3 & 43.6 & 85.5 &33.7\\
        YFCC15M    & CLIP~\cite{radford2021learning} &3.2 & 8.1 & 14.4 & 42.0 & 82.3 &30.0\\
                   & \method  &6.4 & 11.1 & 22.4 & 48.9 & 88.2 &35.4\\
        Merged-30M & CLIP~\cite{radford2021learning} &5.8 & 10.2 & 21.0 & 45.8 & 86.9 &33.9\\
                   & \method  &\bf8.1 & \bf12.5 &\bf 25.3 & \bf49.9 &\bf 90.9 &\bf37.3 \\
        Laion-400M & CLIP~\cite{radford2021learning} &6.1 & 12.2 & 21.3 & 46.3 & 88.3 &34.8\\
    \end{tblr}}
    \vspace{-15pt}
\end{table*}

\subsection{Semantic Segmentation}
To certify the fine-grained representational capacity of \method, we evaluate the transferable performance on semantic segmentation tasks following SAN~\cite{xu2023side}. 
Concretely, We replace and fix the pre-trained model in SAN and fine-tune it with COCO-stuff~\cite{caesar2018coco}. 
As shown in~\Cref{tab:ss}, \method significantly surpasses CLIP across all scale pretraining datasets (\ie, 3M, 12M, 15M and 30M).
\textbf{Notably}, \method exceeds CLIP by 2.5\% on average over 5 widely used semantic segmentation datasets with fewer data (30m \textit{v.s.} 400m).
The results indicate \method can provide reliable fine-grained clues for downstream semantic segmentation tasks with the help of long captions.

\begin{table}[h!]
    \caption{Zero-shot transfer evaluation of different models. Performance on ImageNet and 10 common downstream datasets are reported. We highlight the best performance of each setting in \bf{bold}. Long caption generated from ShareGPT4V~\cite{sharegpt4v} is used.}
    \vspace{-5pt}
    \label{table:zeroshot-main}
    \centering\scriptsize
    \resizebox{\linewidth}{!}{
    \begin{tabular}{c@{\hspace{3.0em}}c@{\hspace{1.0em}}|c@{\hspace{0.7em}}c@{\hspace{0.7em}}c@{\hspace{0.7em}}c@{\hspace{0.7em}}c@{\hspace{0.7em}}c@{\hspace{0.7em}}c@{\hspace{0.7em}}c@{\hspace{0.7em}}c@{\hspace{0.7em}}c@{\hspace{0.7em}}c@{\hspace{0.7em}}|c@{\hspace{0.7em}}}
        \toprule[1.2pt]
        Data&Model&
        \rotatebox[origin=lb]{90}{\smash{ Food-101}} & \rotatebox[origin=lb]{90}{\smash{ CIFAR-10}} & \rotatebox[origin=lb]{90}{\smash{ CIFAR-100}}   & \rotatebox[origin=lb]{90}{\smash{ SUN397}}   & \rotatebox[origin=lb]{90}{\smash{ Cars}}     & \rotatebox[origin=lb]{90}{\smash{ Aircraft}}    & \rotatebox[origin=lb]{90}{\smash{ DTD}}      & \rotatebox[origin=lb]{90}{\smash{ Pets}}     & \rotatebox[origin=lb]{90}{\smash{ Caltech-101}} &
        \rotatebox[origin=lb]{90}{\smash{ Flowers}}  & \rotatebox[origin=lb]{90}{\smash{ Average}}  & \rotatebox[origin=lb]{90}{\smash{ ImageNet}} \\
        \midrule
        \multicolumn{14}{c}{ \textit{Model Architecture: ViT-B/32}}\\
        \midrule
        \multirow{2}{3em}{\rotatebox[origin=c]{0}{\scriptsize{CC3M}}}  &  CLIP~\cite{radford2021learning} & 10.2& 71.3& 32.1& 33.8& 1.4& 1.0& 12.0& 12.1& 50.9& 10.8& 23.6 &  17.2   \\
        &  \method  & 16.1& 82.0& 45.4& 41.3& 2.5& 1.0& 13.9& 18.8& 64.4& 14.1& \bf 30.0 & \bf 25.9 \\
        \midrule
        \multirow{2}{3em}{\rotatebox[origin=c]{0}{\scriptsize{CC12M}}}  &  CLIP~\cite{radford2021learning} &  26.5& 72.5& 38.0& 37.1& 13.7& 2.6& 11.4& 46.2& 74.0& 25.7& 34.8&  32.9  \\
        &  \method  & 48.9& 86.4& 63.0& 55.7& 17.9& 1.9& 23.5& 41.9& 83.2& 25.8& \bf 44.8&  \bf 44.2  \\
        \midrule
        \multirow{9}{3em}{\rotatebox[origin=c]{0}{\scriptsize{YFCC15M}}}  &   SLIP~\cite{mu2022slip}              & 33.3 & 50.7 & 25.5 & 34.7 & 2.8 & 1.7& 14.4 & 23.5 & 59.9  & 49.0 & 29.6 & 34.3  \\
        &  FILIP~\cite{yao2021filip}           & 43.1 & 65.5 & 33.5 & 50.7 & 3.3 & 3.2& 24.3  & 24.1& 68.8  & 52.7 & 36.9 & 39.5 \\
        &  DeCLIP~\cite{li2021supervision}     & 52.5 & 66.7 & 38.7 & 50.3 & 3.8 & 2.1& 27.7 & 33.8 & 74.7& 60.8  & 41.1  & 43.2 \\
        &  HiCLIP~\cite{geng2023hiclip}        & 51.2 & 74.1 & 46.0 & 50.6 & 4.5 & 3.6& 23.1 & 37.8  & 67.4& 60.9 & 41.9 & 40.5 \\
        &  ALIP\cite{yang2023alip}                                & 45.4 & 83.8 & 51.9 & 47.8 & 3.4 & 2.7& 23.2 & 30.7  & 74.1& 54.8 & 41.8 & 40.3  \\
        &  UniCLIP\cite{lee2022uniclip}                             & 48.7 & 78.6 & 47.2 & 50.4 & 3.4 & 2.8 & 23.3 & 32.5 &  73.0  &  8.1 & 36.8 & 42.8  \\
        &  MCD\cite{kim2023misalign}                               & 54.0 & 80.3 & 49.6 & 55.3 & 4.5 & 3.0 & 30.5 & 40.0  &73.2& 7.9   & 39.8 & 44.7  \\
        &  CLIP~\cite{radford2021learning} &26.9& 77.8& 48.2& 42.5& 5.5& 4.7& 18.5& 15.7& 62.0& 39.0& 34.1& 33.3 \\
        &  \method                               & 51.7 &87.9 &60.7 &54.8 &9.4 &7.1 &26.8 &36.3 &79.6 &48.6& \bf 46.3  & \bf 46.6  \\
        \midrule
        \multirow{3}{3em}{\rotatebox[origin=c]{0}{\scriptsize{Merged-30M}}}  &  CLIP~\cite{radford2021learning} & 60.7& 83.6& 54.7& 54.6& 16.2& 6.9& 25.9& 60.5& 81.2& 53.5& 49.8 &49.0\\
        &  HiCLIP\cite{geng2023hiclip} & 63.9 & 77.6 & 56.2  & 60.7 & 22.2 & 5.5 & 38.0&65.6 & 82.4& 62.5  & 53.5 &  52.9  \\
        &  \method &  68.2& 91.8& 69.2& 62.2& 20.7& 8.0& 32.1& 62.8& 86.1& 48.5& \bf 55.0&  \bf 55.7  \\
        \midrule
        \multirow{1}{3em}{\rotatebox[origin=c]{0}{\scriptsize{LAION-400M}}}  &  CLIP~\cite{radford2021learning} & 79.9 & 91.8 & 72.0 & 64.6 & 77.0 & 15.8 & 49.9 & 84.8 & 89.3 & 64.4 & 62.7 & 62.0 \\
        \midrule
        \multicolumn{13}{c}{ \textit{Model Architecture: ViT-B/16}}\\
        \midrule
        \multirow{4}{2.5em}{\rotatebox[origin=c]{0}{\scriptsize CC3M}} & LaCLIP\cite{fan2024improving} & 14.2 & 57.1 & 27.5 & 35.1 & 1.6  & 1.6 & 16.6 & 15.6 & 52.7 & 14.7 & 23.7 & 21.5 \\
        & MLLM-A\cite{liu2023mllms} & 18.7 & 58.4 & 32.4 & 43.8 & 3.9  & 1.5 & 20.2 & 32.1 & 63.5 & 17.5 & 29.2 & 25.0\\
        &  CLIP~\cite{radford2021learning} & 10.6& 53.9& 20.4& 31.2& 1.2& 1.1& 10.4& 11.7& 43.2& 12.9& 19.7&16.0 \\
        & \method                           &  19.4& 74.3& 44.2& 45.9& 2.8& 1.0& 17.0& 27.1& 63.1& 14.7& \bf 31.0 &  \bf 31.1 \\
        \midrule
        \multirow{4}{2.8em}{\rotatebox[origin=c]{0}{\scriptsize CC12M}} 
        & LaCLIP\cite{fan2024improving} & 60.7 & 75.1 & 43.9 & 57.0 & 36.3 & 5.6 & 31.0 & 72.4 & 83.3 & 39.9 & 46.2 & 48.4 \\ 
        & MLLM-A\cite{liu2023mllms} & 60.9 & 83.0 & 55.4 & 59.4 & 24.1 & 3.2 & 30.7 & 64.8 & 79.3 & 36.0 & 49.7 & 47.5 \\
        &  CLIP~\cite{radford2021learning} &25.3& 66.5& 32.1& 39.9& 14.7& 1.9& 13.5& 45.0& 59.8& 15.0& 31.4 &34.0 \\
        & \method & 58.3& 87.3& 62.6& 54.3& 29.7& 4.9& 29.2& 60.3& 83.1& 28.9& \bf 49.9 & \bf 50.3 \\
        \midrule
        \multirow{2}{3em}{\rotatebox[origin=c]{0}{\scriptsize{YFCC15M}}}  &  CLIP~\cite{radford2021learning} & 35.0& 67.1& 34.8& 42.0& 5.1& 6.3& 13.9& 20.4& 54.5& 44.3& 32.3   & 34.1   \\
        &  \method & 44.2& 89.0& 62.0& 57.1& 9.2& 6.4& 30.5& 32.6& 79.8& 40.2& \bf 45.1 &  \bf 48.2  \\
        \midrule
        \multirow{2}{3em}{\rotatebox[origin=c]{0}{\scriptsize{Merged-30M}}}  &  CLIP~\cite{radford2021learning} &64.5& 87.5& 60.3& 61.1& 25.4& 6.9& 33.7& 58.1& 84.5& 57.3& 53.9&  55.2\\
        &  \method &  75.4& 92.3& 70.8& 63.9& 22.7& 7.9& 33.9& 64.1& 88.3& 51.8& \bf 57.1 & \bf 58.4  \\
        \midrule
        \multirow{1}{3em}{\rotatebox[origin=c]{0}{\scriptsize{LAION-400M}}}  &  CLIP~\cite{radford2021learning} & 85.5 & 93.0 & 71.7 & 66.8 & 83.5 & 16.7 & 52.8 & 90.1 & 91.2 & 63.9 & 65.5 & 67.0 \\
        \bottomrule[1.2pt]
    \end{tabular}}
    \vspace{-15pt}
\end{table}

\subsection{Image Recognition}
We have verified the zero-shot classification capability of \method on 11 common classification benchmarks.
Top-1 accuracy is used for evaluation.

For the image classification task, we construct prompts with class label names following the setting in CLIP.
Then text embeddings are extracted from these text inputs with class label names. 
Given an input image, the distances from image embeddings to the text embeddings are computed, and the class label is predicted based on the closest distance. 
We compare \method with Zero-shot CLIP and some state-of-the-art methods on the 11 datasets as mentioned above, demonstrated in~\Cref{table:zeroshot-main}.
\method outperforms other SOTA methods on average over 11 datasets, which approves the ability of our pre-trained model to the downstream tasks. 
It indicates that the long captions are able to enhance the zero-shot performance of CLIP directly transferring to the downstream task.

We also evaluate our model in three multi-classification benchmarks. As shown in~\Cref{tab:multicls}, our method can achieve on par performance with the CLIP trained by 400M datasets. The contents of an image can be so rich that describing them well requires lengthy captions. Thus, an image with lengthy captions can help the model perceive more objects and understand more accurate relations between objects in images.

\begin{table*}[h!]
    \caption{Zero-shot performance on multi-class recognition tasks. OpenImage and ImageNet-Multi contain multiple objects in an image. hico is a relation recognition dataset. We use ViT-B/32 as image backbone. Long caption generated from ShareGPT4V~\cite{sharegpt4v} is used.}
    \vspace{-5pt}
    \label{tab:multicls}
    \centering\scriptsize
    \SetTblrInner{rowsep=1.2pt}      
    \SetTblrInner{colsep=8.0pt}      
    \resizebox{\linewidth}{!}{
    \begin{tblr}{
        cells={halign=c,valign=m},   
        column{1}={halign=l},        
        hline{3,5,7,9}={1-14}{},           
        hline{2}={4-5}{},           
        hline{1,10}={1.0pt},         
        vline{2,3,4,5,6}={1-13}{},      
        cell{1}{1}={r=2}{},          
        cell{1}{2}={r=2}{},          
        cell{1}{3}={r=2}{},          
        cell{1}{4}={c=2}{},          
        cell{1}{6}={r=2}{},          
        cell{3}{1}={r=2}{},          
        cell{5}{1}={r=2}{},          
        cell{7}{1}={r=2}{},          
    }
        Data & Method & ImageNet-Multi & OpenImage    &             & HICO  \\
             &        &                & Class-Common &  Class-Rare & \\
        CC12M      & CLIP~\cite{radford2021learning} &27.69& 58.10	&49.23&13.86  \\
                   & \method  & 37.63&71.48&56.94&19.43 \\
        YFCC15M    & CLIP~\cite{radford2021learning} &26.74& 58.27 &	42.42	&12.25  \\
                   & \method  & 37.85& 66.50 & 50.87 & 18.87 \\
        Merged-30M & CLIP~\cite{radford2021learning} &41.87& 71.94 &	59.01 &	19.83 \\
                   & \method  &48.19 & \bf 74.32 &61.58&\bf 25.36\\
        Laion-400M & CLIP~\cite{radford2021learning} &\bf 53.74& 73.05&\bf 67.44&22.47\\
    \end{tblr}}
\end{table*}

\begin{table*}[h!]
    \caption{Image understanding performance of \method in MLLM. We use ViT-B/32 as image backbone. The best results are in \textbf{bold} and the second best are \underline{underlined}. Long caption generated from ShareGPT4V~\cite{sharegpt4v} is used.}
    \label{tab:mllm}
    \vspace{-5pt}
    \centering\scriptsize
    \SetTblrInner{rowsep=1.2pt}      
    \SetTblrInner{colsep=8.0pt}      
    \resizebox{\linewidth}{!}{
    \begin{tblr}{
        cells={halign=c,valign=m},   
        column{1}={halign=l},        
        hline{2,4}={1-14}{},           
        hline{1,5}={1.0pt},         
        vline{2,3,4,5,6,7}={1-6}{},      
        cell{2}{1}={r=2}{},          
    }
        Data & Method & ScienceQA-IMG &   TextVQA  &     POPE        & MMVP & avg.  \\
        Merged-30M & CLIP~\cite{radford2021learning} & 65.5 & 48.6 & 81.0 & 19.3 & 53.6 \\
                   & \method                         & \underline{66.8} & \underline{48.8} & \textbf{81.8} & \textbf{22.7} & \textbf{55.0} \\
        Laion-400M & CLIP~\cite{radford2021learning} & \textbf{67.3} & \textbf{50.2} & \underline{80.9} & \underline{19.7} & \underline{54.5} \\
    \end{tblr}}
    \vspace{-10pt}
\end{table*}

\subsection{\method in MLLM}
Here we evaluate the image understanding performance of our \method in MLLM. We follow the training process of LLaVA-1.5 \cite{llava1.5} which fixes the visual encoder of CLIP and combines the encoder with LLM, simply replacing LLaVA-1.5's CLIP encoder with several CLIP encoders shown in \cref{tab:mllm} without further tuning. We pick up ScienceQA-IMG \cite{sqa}, TextVQA \cite{Textqa} and POPE \cite{pope} from evaluation benchmarks for LLaVA-1.5 which can be directly obtained the results without submitting for website responses. MMVP benchmark \cite{tong2024eyes} which exposes the visual confusion of CLIP in MLLM is also selected. As demonstrated in \cref{tab:mllm}, our \method trained on 30M datasets outperforms CLIP utilizing the same data scale and achieves a competitive performance with CLIP employing 400M datasets. It indicates that long captions are beneficial for holistic visual understanding through image-text joint training because they augment CLIP-like models to mine the rich visual content hidden in images, and finally enhance the image understanding ability in MLLM.

\subsection{Ablation Studies}

\subsubsection{Effectiveness of Each Component.}
To further explore the effectiveness of short captions, long captions and subcaption-specific grouping loss, we perform ablation experiments based on the zero-shot image-text retrieval task, zero-shot image classification and semantic segmentation.

As shown in \cref{tab:ab_dsm}, we first introduce short captions into the baseline (\ie, CLIP), which achieves similar performance with direct using long captions during training. It demonstrates that long captions and short captions can help learn richer information from image than raw captions. 
Directly training with the long captions may not unleash their potential. Thus, we design a uniform sampling strategy for long captions. In this way, we can observe that the performance is better than directing to use long captions, indicating that model can be benefit from the multi-positive pairs including many sub-captions. 
Then, when combining the short captions and sampling sub-captions together, the performance is further improved. This is because the short captions are concise in describing the whole image, while long captions are more details but has some hallucinations, which can complement each other.
Finally, with the help of grouping loss, our \method can achieve the best performance in terms of all metrics. We design a simple yet effective strategy to use the long captions in language-image pre-training.

\begin{table*}[t]
    \vspace{-5pt}
    \caption{Ablation study of different designs. `S.C' refers to short captions generated by MLLM. `SGL' refers to subcaption-specific grouping losss. We use ViT-B/16 as image backbone.}
    \vspace{-5pt}
    \label{tab:ab_dsm}
    \centering\scriptsize
    \SetTblrInner{rowsep=1.2pt}      
    \SetTblrInner{colsep=1.2pt}      
    \resizebox{\linewidth}{!}{
    \begin{tblr}{
        cells={halign=c,valign=m},   
        hline{4,5,6}={1-14}{},           
        hline{3}={1-2}{},           
        hline{1,10}={1.0pt},         
        vline{2,3,4,5,7,9,10}={1-13}{},      
        cell{1}{5}={c=2}{},          
        cell{1}{7}={c=2}{},          
        cell{1}{3}={r=3}{},          
        cell{1}{4}={r=3}{},          
        cell{1}{1}={r=2,c=2}{},          
    }
         Long Captions&&S.C &SGL & Text Retrieval &          & Image Retrieval &          & Classification &Segmentation  \\
          &&&    & Flickr30k    & MSCOCO  & Flickr30k & MSCOCO  & ImageNet    & VOC-20    \\
          Direct&Sampling&&   & R@1  & R@1   & R@1  & R@1 & Acc.(\%)    &mIOU   \\
        &&&               &32.6&14.8&21.4&11.5&20.3 & 64.4    \\
        &&$\checkmark$&                        &55.1&32.7&43.3&23.0&25.6 & 77.7   \\
        $\checkmark$&&&     & 56.6 &30.2&40.9&21.4&24.4& 75.7 \\
        &$\checkmark$&&     &63.0&35.7&49.0&25.6&30.0&81.8\\
        &$\checkmark$&$\checkmark$&            &68.3&40.8&53.4&29.4&30.1&82.9\\
        &$\checkmark$&$\checkmark$&$\checkmark$&\bf69.5   &\bf42.8&\bf54.4&\bf30.4   &\bf31.1   &\bf84.5
        \end{tblr}}
    \vspace{-5pt}
\end{table*}

\begin{figure}
  \centering
  \vspace{-15pt}
  \begin{subfigure}[b]{0.48\textwidth}
    \includegraphics[width=\linewidth]{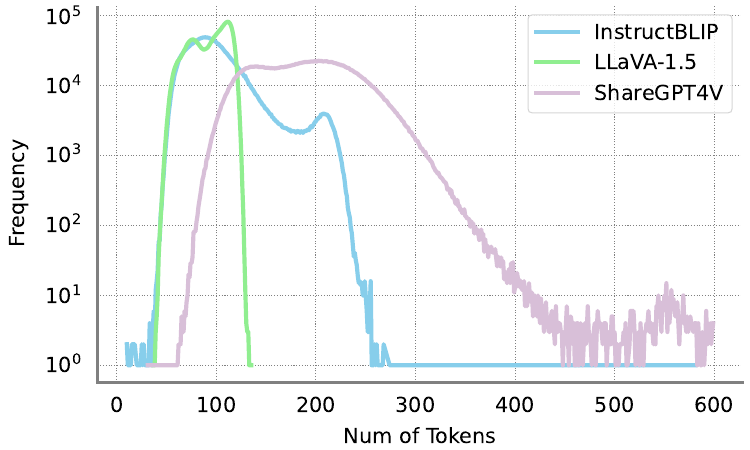}
    \label{fig:sub1}
  \end{subfigure}
  \hfill
  \begin{subfigure}[b]{0.48\textwidth}
    \includegraphics[width=\linewidth]{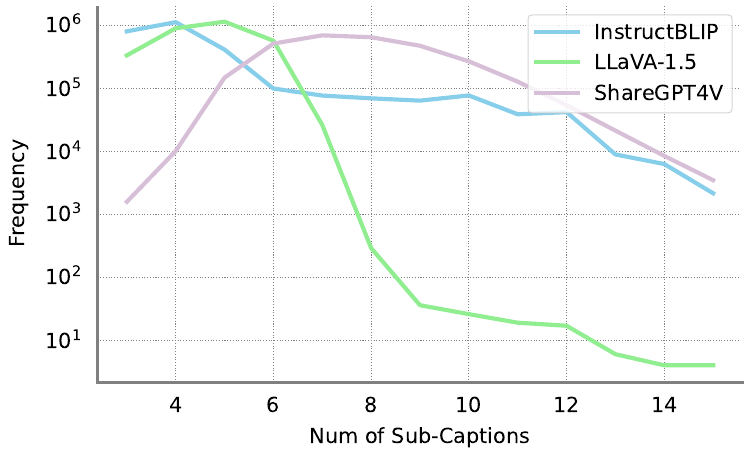}
    \label{fig:sub2}
  \end{subfigure}
  \vspace{-15pt}
  \caption{Statistics of long captions generated by MLLMs (\ie, InstructBLIP, LLAVA-1.5 and ShareGPT4V).}
  \label{fig:both_images}
  \vspace{-15pt}
\end{figure}
\subsubsection{Long Captions from Different MLLMs.}
\label{sec:dMLLM}
Given the significance of synthetic captions in this study, we investigate the impact of captions generated by different MLLMs on downstream tasks.
The experiment results are presented in~\Cref{tab:ab_longcap}.
It is worth noting that using long captions generated by InstructBLIP~\cite{instructblip} achieves better performance than using short captions.
Additionally, we provide some stronger MLLMs to generate long captions, which can bring a more significant improvement. 
As shown in~\Cref{fig:both_images}, detailed captions, characterized by a greater number of tokens and subcaptions, can capture the contents of an image more comprehensively than typically possible with short captions.
Long captions generated by ShareGPT4V~\cite{sharegpt4v} achieve the best performance, and have longer tokens and subcaptions, which demonstrates the effectiveness of its long captions.

\begin{table*}[h!]
    \vspace{-15pt}
    \caption{Ablation study of long captions from different MLLMs. We use ViT-B/16 as image backbone.}
    \label{tab:ab_longcap}
    \centering\scriptsize
    \SetTblrInner{rowsep=1.2pt}      
    \SetTblrInner{colsep=2.4pt}      
    \resizebox{\linewidth}{!}{
    \begin{tblr}{
        cells={halign=c,valign=m},   
        column{1}={halign=l},        
        hline{4,5,8}={1-14}{},           
        hline{1,9}={1.0pt},        
        vline{2,4,6,7}={1-14}{},      
        cell{1}{1}={r=3}{},          
        cell{1}{2}={c=2}{},          
        cell{1}{4}={c=2}{},          
    }
        MLLM & Text Retrieval &      & Image Retrieval &     & Classification & Segmentation\\
             & Flickr30k    & MSCOCO   & Flickr30k    & MSCOCO  & ImageNet   & VOC-20    \\
             & R@1       & R@1    & R@1           & R@1    & Acc.(\%)  & mIOU     \\
        CLIP \textit{w/o} MLLM  & 32.6   &14.8   &21.4   &11.5   &20.3   & 64.4\\
        (1) InstructBLIP &58.7&34.4&45.2&24.9&27.8&79.2\\ 
        (2) LLaVA-1.5&  66.8  & 42.4  & 53.3  & 29.9 &29.0&81.8\\
        (3) ShareGPT4V  & 69.5   &42.8&54.4&30.4   &31.1   &84.5\\
        (1)+(2)+(3) * & 74.4 & 46.4 & 62.4 & 34.9 & 34.6&88.2\\
        \end{tblr}}
    \vspace{-18pt}
\end{table*}

\subsubsection{Number of sub-captions.}

We evaluate the performance for different numbers $K$ of sampled sub-captions from the sub-caption set. As shown in~\Cref{tab:ablation_k}, we observe that as the number of sub-captions increase, the performance gradually improves in terms of zero-shot classification, image-text retrieval, and semantic segmentation. However, when the number $K$ of sub-captions reached approximately 8, the performance showed little to no further improvement. This phenomenon can be attributed to the fact that the number of tokens and sub-captions derived from long captions in the synthetic caption dataset reached its peak, as shown in~\Cref{fig:both_images}. Increasing the number of sub-captions may result in redundant samples, which do not provide additional information to enhance model training.

\begin{table*}[h]
    \caption{Ablation study of sampling number of sub-captions from long captions. We use ViT-B/16 as image backbone.}
    \label{tab:ablation_k}
    \centering\scriptsize
    \SetTblrInner{rowsep=1.2pt}      
    \SetTblrInner{colsep=7.0pt}      
    \resizebox{\linewidth}{!}{
    \begin{tblr}{
        cells={halign=c,valign=m},   
        column{1}={halign=l},        
        hline{4,5}={1-14}{},           
        hline{1,13}={1.0pt},         
        vline{2,4,6,7}={1-14}{},      
        cell{1}{1}={r=3}{},          
        cell{1}{2}={c=2}{},          
        cell{1}{4}={c=2}{},          
    }
        $K$ & Text Retrieval &      & Image Retrieval &     & Classification & Segmentation\\
             & Flickr30k    & MSCOCO   & Flickr30k    & MSCOCO  & ImageNet   & VOC-20    \\
             & R@1       & R@1    & R@1           & R@1    & Acc.(\%)  & mIOU     \\
        CLIP~\cite{radford2021learning} & 32.6   &14.8   &21.4   &11.5   &20.3   & 64.4\\
        3    & 65.4   &37.4   &49.5   &26.9   &29.4   & 82.0\\
        4    & 68.0   &38.2   &51.6   &28.6   &30.8   & 82.1\\
        5    & 68.5   &39.0   &53.6   &29.2   &30.9   & 79.6\\
        6    & 69.1   &40.6   &53.2   &29.5   &\bf 31.3&81.1\\
        7    & 70.0   &41.4   &53.1   &29.6   &31.1   &84.0\\
        8    & \bf70.9&41.5   &53.0   &29.8   &31.0   &84.5\\
        9    & 70.5   &41.9   &54.0   &\bf30.5&31.1   &83.0\\
        10   & 69.5   &\bf42.8&\bf54.4&30.4   &31.1   &\bf 84.5\\
        \end{tblr}}
\end{table*}

\subsection{Visualization}
\subsubsection{Visualization of semantic segmentation and image-text retrieval.} 
\begin{figure*}[t]
    \vspace{-5pt}
    \centering
    \includegraphics[width=\textwidth]{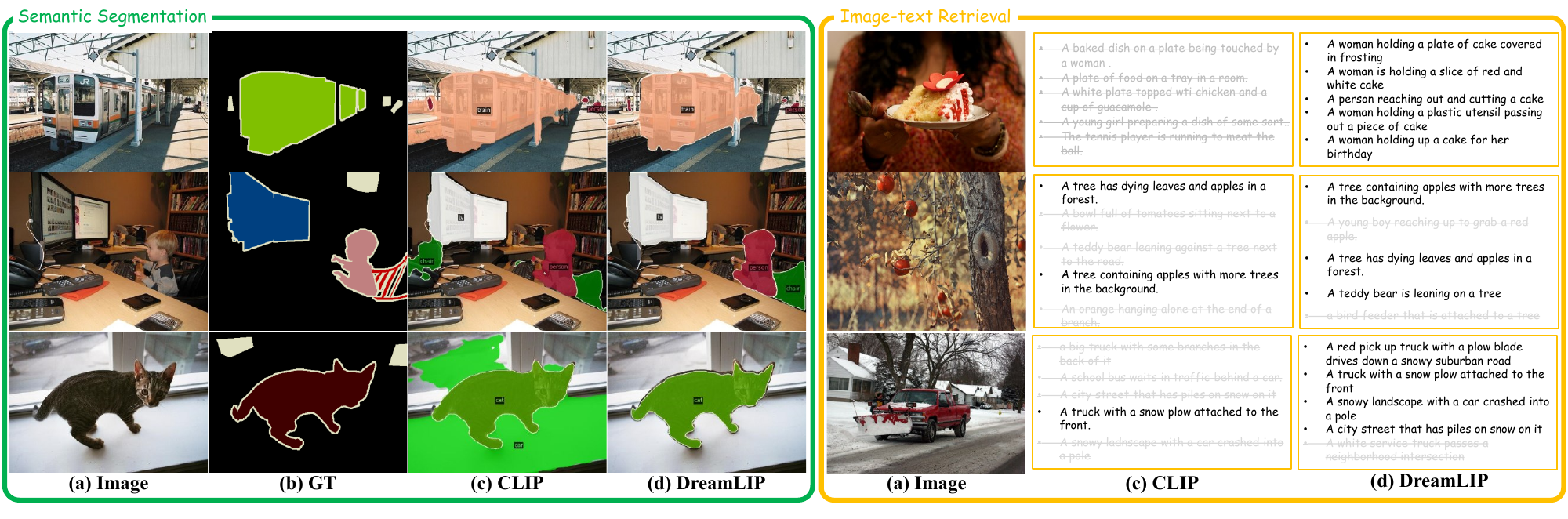}
    \caption{%
        Visualization of semantic segmentation and image-text retrieval.
    }
    \vspace{-5pt}
    \label{fig:vis}
\end{figure*}

In order to offer a comprehensive qualitative understanding, we have curated a set of examples from the VOC and MSCOCO validation sets in~\Cref{fig:vis}, showcasing the notable accuracy improvements achieved by \method. These carefully chosen examples serve as compelling evidence of our method's remarkable ability to effectively distinguish between intricate and nuanced categories. Notably, our method excels in scenarios where vanilla CLIP encounters challenges and struggles to make accurate differentiations. By presenting these examples, we substantiate the claim that our approach significantly enhances the discriminative power of the model, particularly in fine-grained categorization tasks.

\begin{figure}[t]
  \centering
    \vspace{-5pt}
  \includegraphics[width=\textwidth]{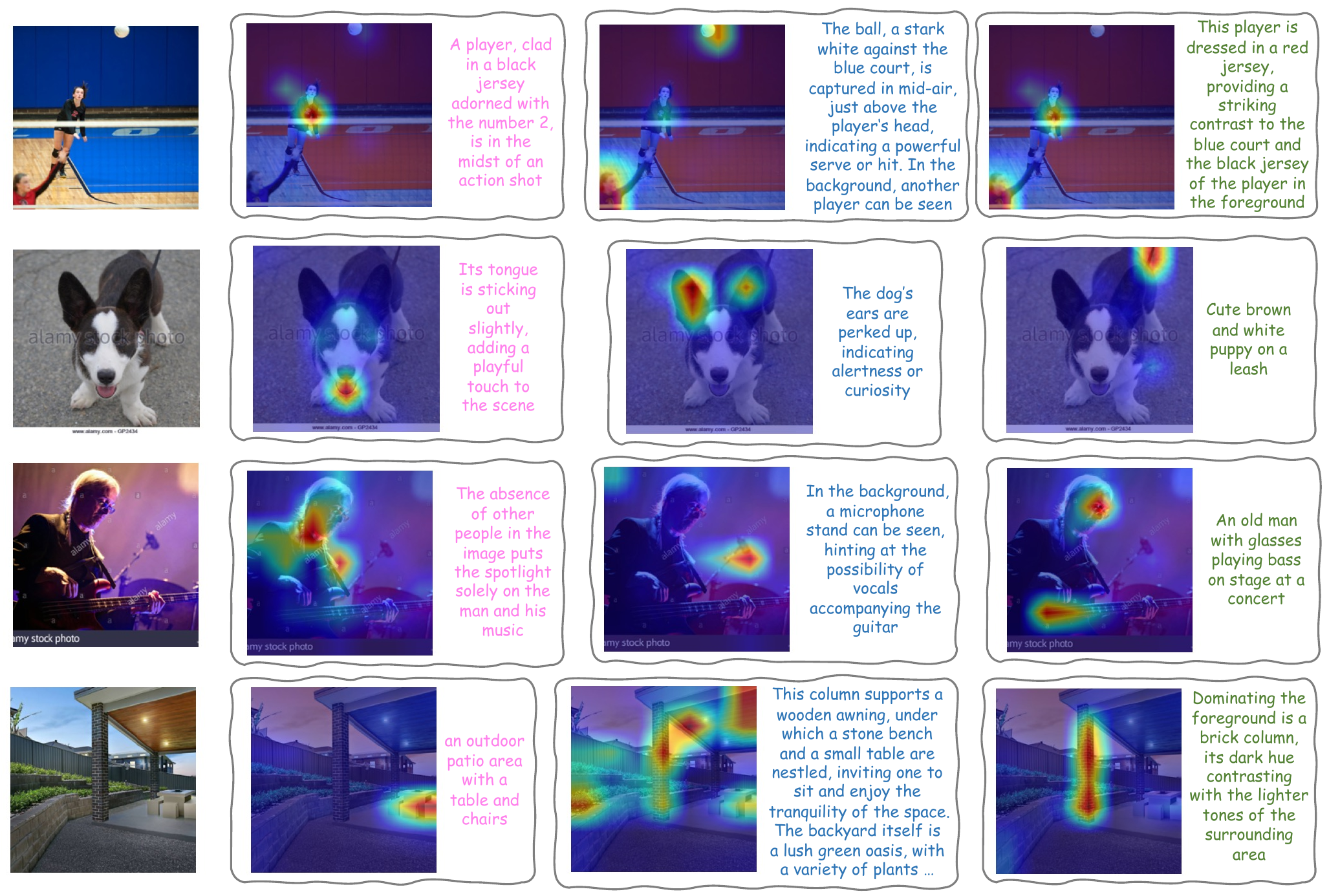}
  \caption{\textbf{Visualization for Attention Map.} The sub-captions corresponding to the attention maps are split from the generated long captions.} 
   \label{visualization} 
    \vspace{-5pt}
\end{figure}

\subsubsection{Attention Map visualization.} We visualize the attention map between different sub-captions from the generated long captions in Fig.\ref{visualization} following \cite{vis1, vis2}. As we motivate above, \method can indeed focus on the corresponding regions according to the different sub-captions. Even the dog's tongue (as shown in row 2 and column 2) and the microphone (as shown in row 3 and column 3) in the noisy background can be precisely perceived by \method.

\section{Conclusion}\label{sec:conclusion}
We re-caption 30M images with detailed descriptions using a pre-trained MLLM and explore the usage of these long captions under a contrastive learning framework.
Specifically, we propose to dynamically sample sub-captions from the text label to construct multiple positive pairs, and introduce a grouping loss to match the embeddings of each sub-caption with its corresponding local image patches in a self-supervised manner.
Experimental results on a wide range of downstream tasks demonstrate the consistent superiority of our method, termed \method, over previous alternatives, highlighting its fine-grained representational capacity.
This work represents a promising direction for enhancing CLIP, and we anticipate it will inspire further research.
%
%
\bibliographystyle{splncs04}
\bibliography{ref.bib}
\clearpage
\appendix
\onecolumn




\section{Long Caption in Multi-modality Learning}

In this section, we discuss various studies related to the generation of extended captions within the context of text-to-image (T2I) synthesis. Several notable works~\cite{chen2023pixart,chen2024pixart,dalle3} have employed Multimodal Large Language Models (MLLM) to produce comprehensive and rich captions for T2I tasks. These approaches let models better capture and draw a picture by utilizing more intricate and accurate captions of scenes. Meanwhile, in language-image pretraining tasks, our objective is for elaborate captions to leverage real-world images more effectively, thereby endowing multimodal foundation models with some additional capabilities (\eg, Vision-Language Compositionality in~\Cref{VLC}).


\section{Sampling from Mixture Generated Long Captions}
\begin{figure}[h]
    \vspace{-10pt}
    \centering
    \includegraphics[width=0.98\linewidth]{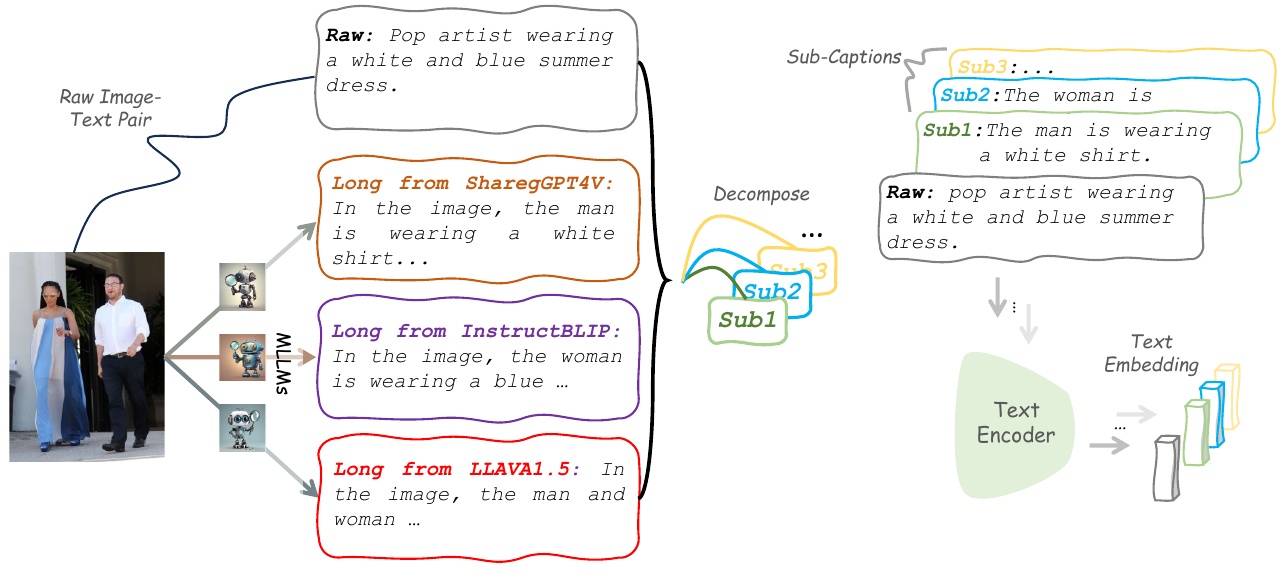}
    \caption{
        Illustration of \method with mixture generated long captions. 
        }
    \label{fig:arch_esb}
    \vspace{-10pt}
\end{figure}

In the main paper, we mainly use the long caption from ShareGPT4V as the training text. However, different MLLMs may focus on different regions of real-world images due to their difference of training process and data. Thus, the generated long/short captions from MLLMs can help each other. Inspired by this straightforward idea, we merge all captions from different MLLMs together, and sample the sub-captions from the set of merged captions. As shown in~\Cref{fig:arch_esb}, we use three kinds of MLLMs to generate the long captions and short captions, and then sample the sub-captions from it as the input of text encoder.

As shown in~\Cref{tab:ablation_km}, the number of sub-captions and tokens exceeds the result of using ShareGPT4V alone. Further, our ablation studies on the mixture generated long captions reveal that an increased number of sub-captions correlates with enhanced performance, surpassing the result achieved with long captions solely from ShareGPT4V. These findings suggest that various MLLMs capture distinct image regions, providing complementary information. 

\begin{figure*}[h]
  \centering
  \begin{subfigure}[b]{0.48\textwidth}
    \includegraphics[width=\linewidth]{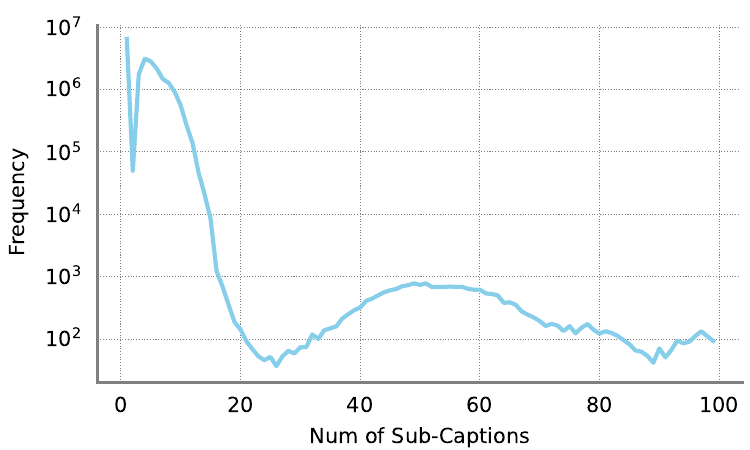}
  \end{subfigure}
  \hfill
  \begin{subfigure}[b]{0.48\textwidth}
    \includegraphics[width=\linewidth]{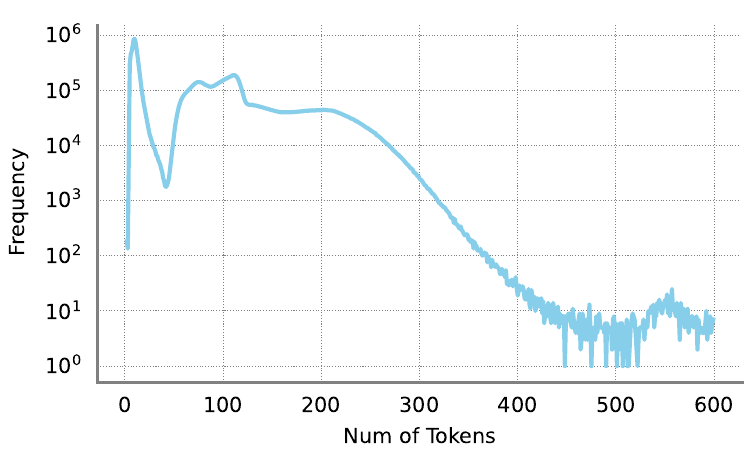}
  \end{subfigure}
  \caption{Statistics of merged caption that include raw caption, three long captions and three short catptions generated by MLLMs (\ie, InstructBLIP, LLAVA-1.5 and ShareGPT4V).}
  \label{fig:slong}
\end{figure*}

We also analyze the mixture generated long captions, characterized by a greater number of tokens and subcaptions, can capture the contents of an image more comprehensively than one kind of captions as shown in~\Cref{fig:slong}. In future work, we aim to investigate the synergistic effects of integrating additional MLLMs.

\begin{table*}[h]
    \vspace{-10pt}
    \caption{Ablation study of sampling number of sub-captions from mixture generated long captions. We use ViT-B/16 as image backbone. `CLIP*' refers to enhancing CLIP with image data augmentation following SimCLR~\cite{chen2020simple}}
    \label{tab:ablation_km}
    \centering\scriptsize
    \SetTblrInner{rowsep=1.2pt}      
    \SetTblrInner{colsep=6.8pt}      
    \resizebox{\linewidth}{!}{
    \begin{tblr}{
        cells={halign=c,valign=m},   
        column{1}={halign=l},        
        hline{4,6}={1-14}{},           
        hline{1,15}={1.0pt},         
        vline{2,4,6,7}={1-14}{},      
        cell{1}{1}={r=3}{},          
        cell{1}{2}={c=2}{},          
        cell{1}{4}={c=2}{},          
    }
        $K$ & Text Retrieval &      & Image Retrieval &     & Classification & Segmentation\\
             & Flickr30k    & MSCOCO   & Flickr30k    & MSCOCO  & ImageNet   & VOC-20    \\
             & R@1       & R@1    & R@1           & R@1    & Acc.(\%)  & mIOU     \\
        CLIP & 29.4 & 14.3 & 19.9 & 10.2 & 16.0 & 62.7 \\
        CLIP* & 32.6 & 14.8 & 21.4 & 11.5 & 20.3 & 64.4\\
        2    & 68.1 & 39.6 & 54.3 & 29.1 & 30.7& 83.1\\
        4    & 72.6 & 44.1 & 58.2 & 32.6 & 30.1&85.5\\
        6    & 74.9 & 44.9 & 59.2 & 33.4 & 31.9&85.4\\
        8    & 75.0 & 45.6 & 60.8 & 33.8 & 32.8&86.6\\
        10   & 74.3 & 47.2 & 61.6 & 34.8 & 33.6&87.4\\
        12   & 75.5 & 46.4 & 61.5 & 35.0 & 34.3&86.7\\
        14   & 75.9 & 47.4 & 61.8 & 35.0 & 33.9&87.2\\
        16   & 75.3 & 48.1 & 61.7 & 35.1 & 34.7&87.8\\
        18   & 74.4 & 46.4 & 62.4 & 34.9 & 34.6&88.2\\
        \end{tblr}}
    \vspace{-10pt}
\end{table*}

\section{Data Augmentation with Synthetic Images Generated by SDXL-turbo}
\begin{figure}[h!]
    \centering
    \includegraphics[width=0.98\linewidth]{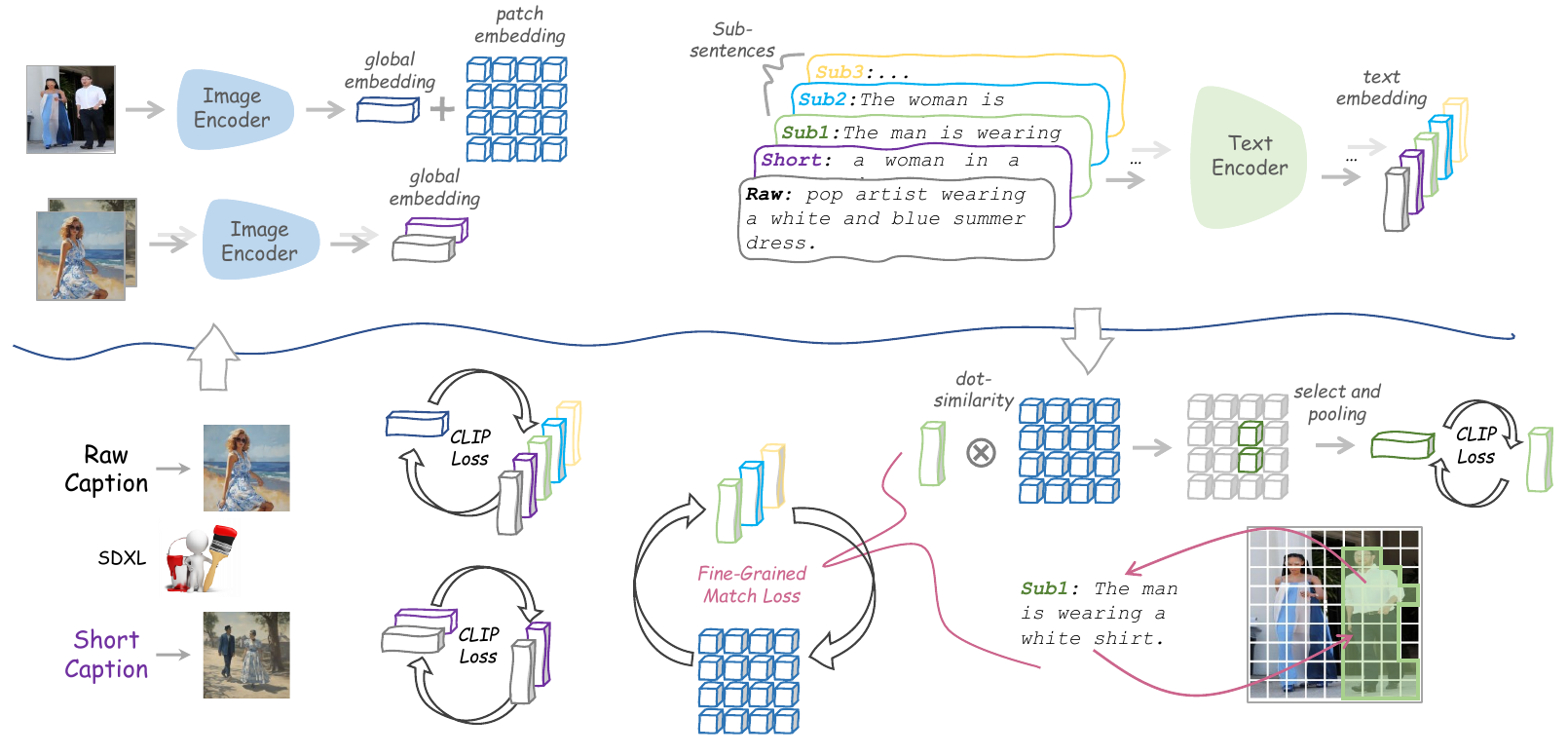}
    \vspace{-2mm}
    \caption{
        Illustration of \method with synthetic images generated by SDXL-turbo. 
        }
    \label{fig:wsdxl}
    \vspace{-5mm}
\end{figure}

Language-image pre-training could benefit from long captions generation, due to the strong captioning capacity of image-to-text models. Meanwhile, as shown in~\Cref{fig:wsdxl}, we would explore whether text-to-image models (\eg, SDXL-turbo~\cite{sdxlturbo}) can bring performance improvement for language-image pre-training. An image may be depicted through a variety of sentences, while a single caption also has the capacity to evoke numerous images. Thus, we adopt the SDXL-turbo to generate some images from raw captions and short captions.
As shown in~\Cref{table:zeroshot-TI}, the DreamLIP with synthetic images outperforms DreamLIP on three kinds of datasets, which approves the ability of the introduction of synthetic images.

\begin{table*}[h!]
    \caption{Zero-shot transfer evaluation of different models. Performance on ImageNet and 10 common downstream datasets are reported.}
    \label{table:zeroshot-TI}
    \centering\scriptsize
    \begin{tabular}{c@{\hspace{2.0em}}c@{\hspace{1.0em}}|c@{\hspace{0.45em}}c@{\hspace{0.45em}}c@{\hspace{0.45em}}c@{\hspace{0.45em}}c@{\hspace{0.45em}}c@{\hspace{0.45em}}c@{\hspace{0.45em}}c@{\hspace{0.45em}}c@{\hspace{0.45em}}c@{\hspace{0.45em}}c@{\hspace{0.45em}}|c@{\hspace{0.4em}}}
        \toprule[1.2pt]
        Data&Model&
        \rotatebox[origin=lb]{90}{\smash{ Food-101}} & \rotatebox[origin=lb]{90}{\smash{ CIFAR-10}} & \rotatebox[origin=lb]{90}{\smash{ CIFAR-100}}   & \rotatebox[origin=lb]{90}{\smash{ SUN397}}   & \rotatebox[origin=lb]{90}{\smash{ Cars}}     & \rotatebox[origin=lb]{90}{\smash{ Aircraft}}    & \rotatebox[origin=lb]{90}{\smash{ DTD}}      & \rotatebox[origin=lb]{90}{\smash{ Pets}}     & \rotatebox[origin=lb]{90}{\smash{ Caltech-101}} &
        \rotatebox[origin=lb]{90}{\smash{ Flowers}}  & \rotatebox[origin=lb]{90}{\smash{ Average}}  & \rotatebox[origin=lb]{90}{\smash{ ImageNet}} \\
        \midrule
        \multicolumn{14}{c}{ \textit{Model Architecture: ViT-B/32}}\\
        \midrule
        \multirow{3}{3em}{\rotatebox[origin=c]{0}{\scriptsize{CC3M}}}  &  CLIP & 10.2& 71.3& 32.1& 33.8& 1.4& 1.0& 12.0& 12.1& 50.9& 10.8& 23.6 &  17.2   \\
        &  \method  & 16.1& 82.0& 45.4& 41.3& 2.5& 1.0& 13.9& 18.8& 64.4& 14.1& \bf 30.0 & \bf 25.9 \\
        &  + Synthetic Images    & 21.7& 80.9& 51.2& 45.1& 3.2& 1.5& 20.8& 24.1& 68.2& 15.6& \bf 33.2 & \bf 29.4  \\
        \midrule
        \multirow{3}{3em}{\rotatebox[origin=c]{0}{\scriptsize{CC12M}}}  &  CLIP &  26.5& 72.5& 38.0& 37.1& 13.7& 2.6& 11.4& 46.2& 74.0& 25.7& 34.8&  32.9  \\
        &  \method  & 48.9& 86.4& 63.0& 55.7& 17.9& 1.9& 23.5& 41.9& 83.2& 25.8& \bf 44.8&  \bf 44.2  \\
        & + Synthetic Images   & 46.0& 86.1& 57.2& 53.3& 26.1& 3.3& 26.3& 56.4& 83.7& 31.3& \bf 47.0  & \bf 46.4  \\
        \midrule
        \multirow{3}{3em}{\rotatebox[origin=c]{0}{\scriptsize{YFCC15M}}}  & CLIP &26.9& 77.8& 48.2& 42.5& 5.5& 4.7& 18.5& 15.7& 62.0& 39.0& 34.1& 33.3 \\
        &  \method                               & 51.7 &87.9 &60.7 &54.8 &9.4 &7.1 &26.8 &36.3 &79.6 &48.6& \bf 46.3  & \bf 46.6  \\
        &  + Synthetic Images                              & 54.2& 87.1& 57.9& 53.5& 14.1& 8.7& 31.4& 32.3& 80.8& 40.5& \bf 46.1 & \bf 47.5\\
        \midrule
        \multicolumn{13}{c}{ \textit{Model Architecture: ViT-B/16}}\\
        \midrule
        \multirow{3}{2.5em}{\rotatebox[origin=c]{0}{\scriptsize CC3M}} 
        &  CLIP &10.3 & 54.9 & 21.8 & 25.0 & 0.8  & 1.4 & 10.5 & 12.8 & 43.3 & 10.2 & 19.1&20.3\\
        & \method                           &  19.4& 74.3& 44.2& 45.9& 2.8& 1.0& 17.0& 27.1& 63.1& 14.7& \bf 31.0 &  \bf 31.1 \\
        &   + Synthetic Images                          &  22.8& 72.8& 43.0& 46.6& 3.9& 1.2& 22.1& 25.7& 70.0& 17.7& \bf 32.6 &  \bf 33.3 \\
        \midrule
        \multirow{3}{2.8em}{\rotatebox[origin=c]{0}{\scriptsize CC12M}} 
        &  CLIP &25.3& 66.5& 32.1& 39.9& 14.7& 1.9& 13.5& 45.0& 59.8& 15.0& 31.4 &34.0 \\
        & \method & 58.3& 87.3& 62.6& 54.3& 29.7& 4.9& 29.2& 60.3& 83.1& 28.9& \bf 49.9 & \bf 50.3 \\
        &   + Synthetic Images  & 58.3& 87.3& 64.6& 53.9& 29.7& 4.9& 29.2& 60.3& 83.1& 28.9& \bf 50.0 & \bf 50.7 \\
        \midrule
        \multirow{2}{3em}{\rotatebox[origin=c]{0}{\scriptsize{YFCC15M}}}  &  CLIP & 35.0& 67.1& 34.8& 42.0& 5.1& 6.3& 13.9& 20.4& 54.5& 44.3& 32.3   & 34.1   \\
        &  \method & 44.2& 89.0& 62.0& 57.1& 9.2& 6.4& 30.5& 32.6& 79.8& 40.2& \bf 45.1 &  \bf 48.2  \\
        \bottomrule[1.2pt]
    \end{tabular}
\end{table*}

\section{Evaluation on Vision-Language Compositionality}
\label{VLC}
We compare DreamLIP with CLIP on Attribution, relation, ordering (ARO)~\cite{yuksekgonul2022and} and SugarCrepe~\cite{hsieh2024sugarcrepe} benchmark. 
These two benchmarks are used to measure compositional understanding of vision-language models. 
Results are shown in~\Cref{tab:relation_aro_sugar}.
DreamLIP significantly outperforms CLIP across all tasks when pretrained on the same dataset (Merged-30M).
Also, DreamLIP-30M achieves better results on 7 out of 11 tasks compared to CLIP-400M.
It indicates the usage of long captions with detailed descriptions well enhance model's compositional understanding.

\begin{table*}[h!]
    \caption{Results on the ARO~\cite{yuksekgonul2022and} and SugarCrepe~\cite{hsieh2024sugarcrepe} benchmark. 
    CLIP-30M and -400M indicate the CLIP is pre-trained on Merged-30M and LAION-400M dataset, respectively. DreamLIP-30M is pre-trained on Merged-30M dataset}
    \label{tab:relation_aro_sugar}
    \centering\scriptsize
    \SetTblrInner{rowsep=0.1pt}      
    \SetTblrInner{colsep=0.8pt}      
    \resizebox{\linewidth}{!}{
    \begin{tblr}{
        cells={halign=c,valign=m},   
        column{1}={halign=l},        
        hline{2,4,6}={1-14}{},           
        hline{1,7}={1.0pt},         
        vline{2,6}={1-14}{},
        vline{4,5,9,11}={1-14}{},      
        cell{1}{1}={r=3}{},          
        cell{1}{2}={c=4}{},          
        cell{1}{6}={c=7}{},          
        cell{2}{2}={c=2}{},          
        cell{2}{6}={c=3}{},          
        cell{2}{9}={c=2}{},          
        cell{2}{11}={c=2}{},          
    }
        Model & \bf Aro &  &     &        & \bf SugarCrepe &   & && &&   \\
              & VG  &  & COCO& Flickr &Replace &&&Swap&&Add& \\
              & Attribution  & Relation & Order& Order &Object &Attribute&Relation&Object&Attribute&Object& Attribute\\
        CLIP-30M & 58.28 &43.52 & 23.42 & 27.02 & 82.99 & 73.10&59.25 & 60.00 & 65.17 & 69.84 &63.15\\
        DreamLIP-30M & \bf78.17 & \bf53.62 & \bf41.26 &\bf42.78 & 91.46 &82.23&\bf72.40 &\bf69.80 &\bf79.43 & 81.38 &77.46\\
        CLIP-400M & 59.92 & 47.86 & 38.83 & 42.56 & \bf91.58 &\bf82.75 &67.57 & 61.22& 68.62 & \bf82.01 &\bf78.61\\
             
        \end{tblr}
        }
\end{table*}

\section{Experiments}

\subsection{Hyper-Parameters}

Table~\ref{table:hyperparam} provides an overview of the pre-training hyperparameters used for CLIP on all datasets.  
Further details can be found in Table~\ref{table:hyperparam}. The pre-training processes of CC12M, YFCC15M and Merged-30M were conducted on four machines with eight A100 GPUs. For CC3M, eight A100 GPUs are used.

\begin{table*}[h!]
\centering
\caption{
\small Detailed pre-training hyper-parameters for CLIP training on all four image-text datasets.}
\label{table:hyperparam}

\subfloat[
\small
Hyper-parameter on CC3M.
\label{table:hyperparam-cc3m}
]{
\centering
\begin{minipage}{0.47\linewidth}{\begin{center}
\resizebox{0.92\textwidth}{!}{
\begin{tabular}{l|l}
\toprule
Config & Value \\
\midrule
Batch size & $1,024$ \\
Optimizer & AdamW~\cite{loshchilov2017decoupled} \\
Learning rate & $5\times10^{-4}$ \\
Weight decay & $0.5$ \\
Adam $\beta$ & $\beta_1, \beta_2=(0.9, 0.98)$\\
Adam $\epsilon$ & $1\times10^{-8}$ \\
Total epochs & $32$ \\
Warm up iterations & $2,000$\\
Learning rate schedule & cosine decay \\
\bottomrule
\end{tabular}}
\end{center}}\end{minipage}
}
\subfloat[
\small
Hyper-parameter on CC12M.
\label{table:hyperparam-cc12m}
]{
\centering
\begin{minipage}{0.47\linewidth}{\begin{center}
\resizebox{0.92\textwidth}{!}{
\begin{tabular}{l|l}
\toprule
Config & Value \\
\midrule
Batch size & $8,192$ \\
Optimizer & AdamW~\cite{loshchilov2017decoupled} \\
Learning rate & $5\times10^{-4}$ \\
Weight decay & $0.5$ \\
Adam $\beta$ & $\beta_1, \beta_2=(0.9, 0.98)$\\
Adam $\epsilon$ & $1\times10^{-8}$ \\
Total epochs & $32$ \\
Warm up iterations & $2,000$\\
Learning rate schedule & cosine decay \\
\bottomrule
\end{tabular}}
\end{center}}\end{minipage}
}
\\
\subfloat[
\small
Hyper-parameter on YFCC15M.
\label{table:hyperparam-redcaps}
]{
\centering
\begin{minipage}{0.47\linewidth}{\begin{center}
\resizebox{0.92\textwidth}{!}{
\begin{tabular}{l|l}
\toprule
Config & Value \\
\midrule
Batch size & $8,192$ \\
Optimizer & AdamW~\cite{loshchilov2017decoupled} \\
Learning rate & $5\times10^{-4}$ \\
Weight decay & $0.5$ \\
Adam $\beta$ & $\beta_1, \beta_2=(0.9, 0.98)$\\
Adam $\epsilon$ & $1\times10^{-8}$ \\
Total epochs & $32$ \\
Warm up iterations & $2,000$\\
Learning rate schedule & cosine decay \\
\bottomrule
\end{tabular}}
\end{center}}\end{minipage}
}
\subfloat[
\small
Hyper-parameter on Merged-30M.
\label{table:hyperparam-laion}
]{
\centering
\begin{minipage}{0.47\linewidth}{\begin{center}
\resizebox{0.92\textwidth}{!}{
\begin{tabular}{l|l}
\toprule
Config & Value \\
\midrule
Batch size & $8,192$ \\
Optimizer & AdamW~\cite{loshchilov2017decoupled} \\
Learning rate & $5\times10^{-4}$ \\
Weight decay & $0.2$ \\
Adam $\beta$ & $\beta_1, \beta_2=(0.9, 0.98)$\\
Adam $\epsilon$ & $1\times10^{-6}$ \\
Total epochs & $32$ \\
Warm up iterations & $2,000$ \\
Learning rate schedule & cosine decay \\
\bottomrule
\end{tabular}}
\end{center}}\end{minipage}
}
\\
\end{table*}

\subsection{Additional Ablation study}

\subsubsection{Different Image Backbones in semantic segmentation.}
\Cref{tab:ss_diff_backbone} shows the transferable performance of CLIP and DreamLIP with different image backbones on semantic segmentation tasks. 
DreamLIP always achieves better performance than CLIP across different pre-training data and different image backbones.

\begin{table*}[h!]
    \caption{Transferable performance of semantic segmentation on ADE-847, PC-459, ADE-150, PC-59, and VOC-20. Following SAN~\cite{xu2023side}, we used the full training set of COCO-stuff as the training data and our \method as pretrained models.}
    \label{tab:ss_diff_backbone}
    \centering\scriptsize
    \SetTblrInner{rowsep=0.3pt}      
    \SetTblrInner{colsep=6.8pt}      
    \begin{tblr}{
        cells={halign=c,valign=m},   
        column{1}={halign=l},        
        cell{2}{1}={halign=c}, 
        cell{12}{1}={halign=c}, 
        hline{2,3,5,7,9,11,12,13,15,17,19,21}={1-8}{},           
        hline{1,22}={1-8}{1.0pt},         
        vline{2,3,4,5,6,7,8}={1-21}{},      
        cell{2}{1}={c=7}{},          
        cell{12}{1}={c=7}{},          
        cell{3}{1}={r=2}{},          
        cell{5}{1}={r=2}{},          
        cell{7}{1}={r=2}{},          
        cell{9}{1}={r=2}{},          
        cell{13}{1}={r=2}{},          
        cell{15}{1}={r=2}{},          
        cell{17}{1}={r=2}{},          
        cell{19}{1}={r=2}{},          
    }
        Data & Method & ADE-847& PC-459 &ADE-150 &PC-59& VOC-20&avg. \\
        \textit{Model Architecture: ViT-B/32}& &&&&&& \\
        CC3M       & CLIP &2.1 & 5.2 & 12.3 & 33.8 & 65.4 &23.8\\
                   & \method  &4.1 & 7.5 & 17.1 & 39.9 & 76.5&29.0 \\
        CC12M      & CLIP &3.3 & 6.7 & 15.7 & 39.2 & 79.7&28.9 \\
                   & \method  &6.1 & 10.0 & 23.3 & 43.6 & 85.5 &33.7\\
        YFCC15M    & CLIP &3.2 & 8.1 & 14.4 & 42.0 & 82.3 &30.0\\
                   & \method  &6.4 & 11.1 & 22.4 & 48.9 & 88.2 &35.4\\
        Merged-30M & CLIP &5.8 & 10.2 & 21.0 & 45.8 & 86.9 &33.9\\
                   & \method  &\bf8.1 & \bf12.5 &\bf 25.3 & \bf49.9 &\bf 90.9 &\bf37.3 \\
        Laion-400M & CLIP &6.1 & 12.2 & 21.3 & 46.3 & 88.3 &34.8\\
        \textit{Model Architecture: ViT-B/16}& &&&&& &\\
        CC3M       & CLIP &1.9 & 5.3 & 11.4 & 34.5 & 64.4 &23.5\\
                   & \method  &4.9 & 8.7 & 20.5 & 45.0 & 84.5&32.7 \\
        CC12M      & CLIP &3.4 & 7.9 & 16.4 & 39.5 & 80.4 &29.5\\
                   & \method  &6.6 & 12.3 & 23.7 & 48.4 & 85.2 & 35.2\\
        YFCC15M    & CLIP &1.2 & 4.9 & 13.9 & 41.5 & 74.0 &27.1\\
                   & \method  &6.6 & 13.5 & 24.7 & 51.4 & 90.9&37.4 \\
        Merged-30M & CLIP &7.3 & 12.1 & 25.6 & 49.1 & 86.4&36.1 \\
                   & \method  &9.8 & \bf15.4 &\bf 30.6 &\bf 55.1 & 92.2 &\bf 40.6 \\
        Laion-400M & CLIP &\bf10.1 & 12.6 & 27.5 & 53.8 &\bf 94.0 &\bf40.6\\
    \end{tblr}
\end{table*}

\subsubsection{Ablation study of $\sigma$.}
\Cref{tab:sigma} shows the performance of DreamLIP when adjusting $\sigma$. $\sigma$ controls the sparsity of subcaption-specific grouping visual tokens, as shown in Eq.6 in the main paper. Larger $\sigma$ results in sparser subcaption-specific grouping visual tokens within an image.

\begin{table*}[h!]
    \caption{Ablation study of $\sigma$. $\sigma$ is a sparsity threshold as shown in Eq.6 in the main paper. It controls the sparsity of subcaption-specific grouping visual tokens. ViT-B/16 is used as the image backbone. All models are pretrained on CC3M.}
    \label{tab:sigma}
    \centering\scriptsize
    \SetTblrInner{rowsep=1.2pt}      
    \SetTblrInner{colsep=6.8pt}      
    \resizebox{\linewidth}{!}{
    \begin{tblr}{
        cells={halign=c,valign=m},   
        column{1}={halign=l},        
        hline{4,6}={1-14}{},           
        hline{1,11}={1.0pt},         
        vline{2,4,6,7}={1-14}{},      
        cell{1}{1}={r=3}{},          
        cell{1}{2}={c=2}{},          
        cell{1}{4}={c=2}{},          
    }
        $\sigma$ & Text Retrieval &      & Image Retrieval &     & Classification & Segmentation\\
             & Flickr30k    & MSCOCO   & Flickr30k    & MSCOCO  & ImageNet   & VOC-20    \\
             & R@1       & R@1    & R@1           & R@1    & Acc.(\%)  & mIOU     \\
        CLIP & 29.4 & 14.3 & 19.9 & 10.2 & 16.0 & 62.7 \\
        CLIP* & 32.6   &14.8   &21.4   &11.5   &20.3   & 64.4\\
        0.0   & 69.5 & 42.8 & 54.4 & 30.4 & 31.1&84.5\\
        0.1   & 71.0 & 41.2 & 54.3 & 30.4 & 31.9&84.2\\
        0.3   & 71.1 & 41.6 & 54.2 & 30.3 & 31.7&84.0\\
        0.5   & 70.9 & 42.4 & 54.4 & 30.0 & 31.7&85.0\\
        0.7   & 72.5 & 42.2 & 54.9 & 30.4 & 31.8&85.8\\
        \end{tblr}}
\end{table*}

\subsubsection{Ablation study of $\lambda_{S}$.}
We present the influence of $\lambda_{S}$ in~\Cref{tab:lambda}. $\lambda_{S}$ denotes the weight of fine-grained alignment contrastive loss, as shown in Eq.9 in the main paper. 
For most global-level tasks (retrieval and classification), the best performance is reached when $\lambda_{S}=0.7$. 
For local-level tasks, \ie,  semantic segmentation, the best performance is reached when $\lambda_{S}=0.9$. 
The results indicate the fine-grained alignment contrastive loss is helpful for vision-language alignment. 
Larger $\lambda_{S}$ leads the model to learn more fine-grained clues.

\begin{table*}[h!]
    \caption{Ablation study of $\lambda_{S}$, which controls the weight of fine-grained alignment contrastive loss.
     ViT-B/16 is used as image backbone. All models are pretrained on CC3M.}
    \label{tab:lambda}
    \centering\scriptsize
    \SetTblrInner{rowsep=1.2pt}      
    \SetTblrInner{colsep=6.8pt}      
    \resizebox{\linewidth}{!}{
    \begin{tblr}{
        cells={halign=c,valign=m},   
        column{1}={halign=l},        
        hline{4,6}={1-14}{},           
        hline{1,11}={1.0pt},         
        vline{2,4,6,7}={1-14}{},      
        cell{1}{1}={r=3}{},          
        cell{1}{2}={c=2}{},          
        cell{1}{4}={c=2}{},          
    }
         $\lambda_{S}$ & Text Retrieval &      & Image Retrieval &     & Classification & Segmentation\\
             & Flickr30k    & MSCOCO   & Flickr30k    & MSCOCO  & ImageNet   & VOC-20    \\
             & R@1       & R@1    & R@1           & R@1    & Acc.(\%)  & mIOU     \\
        CLIP & 29.4 & 14.3 & 19.9 & 10.2 & 16.0 & 62.7 \\
        CLIP* & 32.6   &14.8   &21.4   &11.5   &20.3   & 64.4\\
        0.1                 &69.5 & 42.8 & 54.4 & 30.4 & 31.1& 84.5\\
        0.3                 &71.0 & 41.2 & 54.3 & 30.4&31.9&84.2\\
        0.5                 &70.2 & 42.2 & 55.0 & 30.8&31.7&83.6\\
        0.7                 &71.4 & 42.6 & 55.6 & 31.6&32.1&84.6\\
        0.9                 &71.1 & 42.2 & 55.9 & 31.5&31.9&85.0\\
        \end{tblr}}
\end{table*}

\begin{figure*}[h!]
    \centering 
    \begin{subfigure}[b]{0.32\linewidth}
         \centering
         \includegraphics[width=\linewidth]{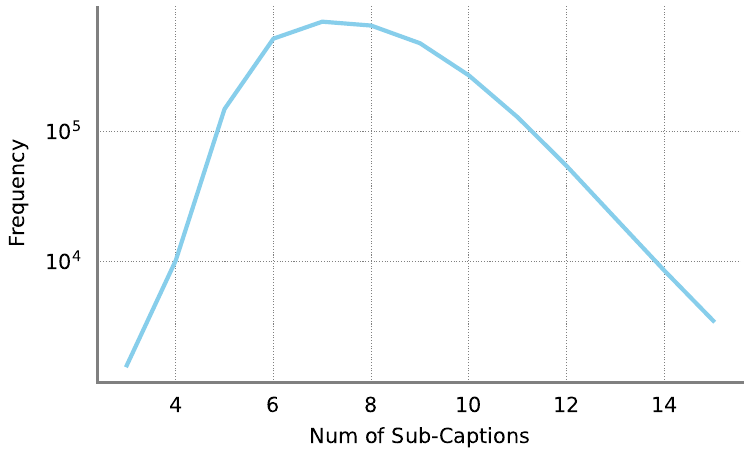}
         \caption{CC3M}
         \label{fig:cc3m_cap_num}
    \end{subfigure}
    \begin{subfigure}[b]{0.32\linewidth}
         \centering
         \includegraphics[width=\linewidth]{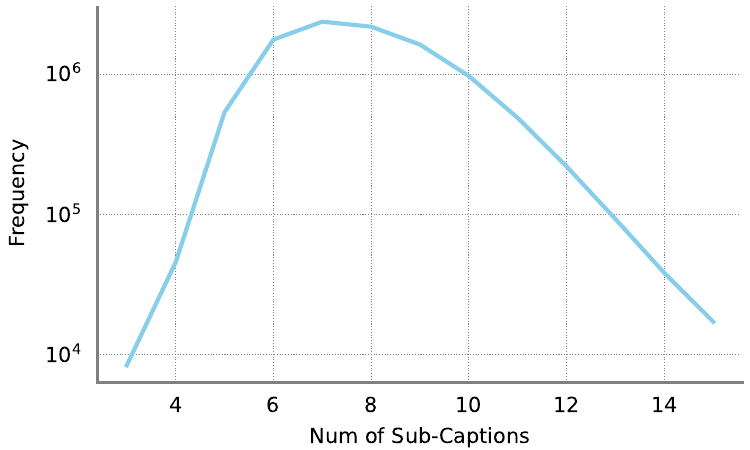}
         \caption{CC12M}
         \label{fig:cc12m_cap_num}
    \end{subfigure}
    \begin{subfigure}[b]{0.32\linewidth}
         \centering
         \includegraphics[width=\linewidth]{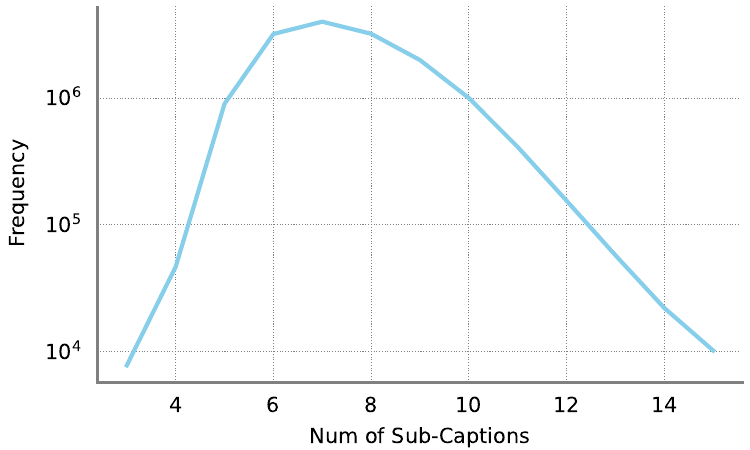}
         \caption{YFCC15M}
         \label{fig:yfcc15m_cap_num}
    \end{subfigure}
    
    \vspace{-2pt}
    \begin{subfigure}[b]{0.32\linewidth}
         \centering
         \includegraphics[width=\linewidth]{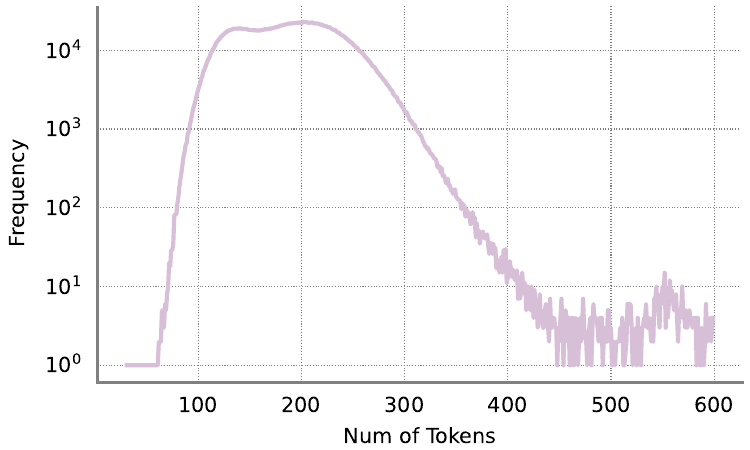}
         \caption{CC3M}
         \label{fig:cc3m_token_num}
    \end{subfigure}
    \begin{subfigure}[b]{0.32\linewidth}
         \centering
         \includegraphics[width=\linewidth]{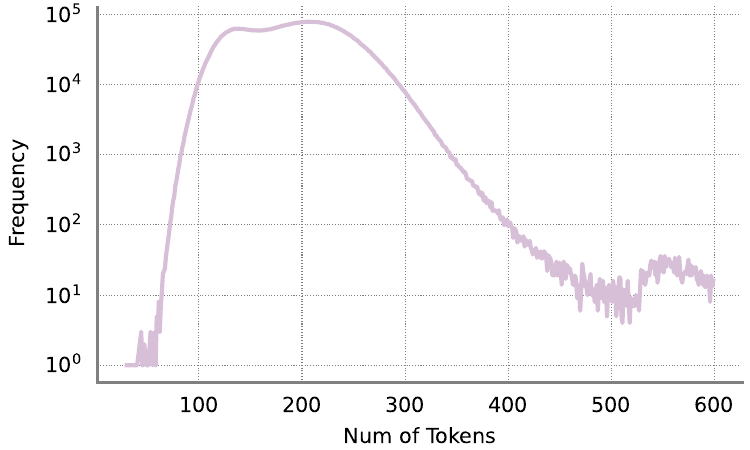}
         \caption{CC12M}
         \label{fig:cc12m_token_num}
    \end{subfigure}
    \begin{subfigure}[b]{0.32\linewidth}
         \centering
         \includegraphics[width=\linewidth]{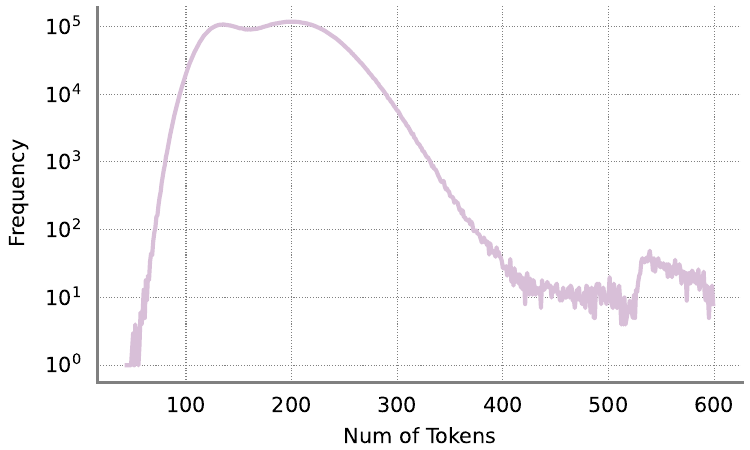}
         \caption{YFCC15M}
         \label{fig:yfcc15m_token_num}
    \end{subfigure}
    \vspace{-.1in}
    \caption{Some statistics of long captions generated by ShareGPT4V. (a)-(c) refer to number of sub-captions from long captions; (d)-(f) refer to number of tokens in long captions.}
    \label{fig:caption_sta} 
    \vspace{-10pt}
\end{figure*}

\subsection{Statistic of Long Captions on Different Datasets}
We conducted some statistics (\ie, the number of tokens and sub-captions) of long captions generated by ShareGPT4V on different datasets. (a)-(c) refer to number of sub-captions from long captions; (d)-(f) refer to number of tokens in long captions.

\section{Limitations}
The existing multimodal large models suffer from hallucinations, with longer captions leading to more severe hallucinations. Directly using the generated long captions will introduce much noise. How to solve the multimodal hallucination problem under long captions, which can further improve the performance of our method.

\end{document}